\documentclass[preprint, 12pt]{elsarticle}
\makeatletter
\def\ps@pprintTitle{%
 \let\@oddhead\@empty
 \let\@evenhead\@empty
 \def\@oddfoot{\centerline{\thepage}}%
 \let\@evenfoot\@oddfoot}
\makeatother

\let\citep\cite

\usepackage{lineno}
\usepackage{amsmath}
\usepackage{amssymb}
\usepackage[autostyle=true]{csquotes}
\usepackage{url}
\usepackage{multirow}
\usepackage{booktabs}
\usepackage{subcaption}
\usepackage{hyperref}

\usepackage{graphicx}
\hypersetup{
    colorlinks=true,
    linkcolor=blue,
    filecolor=magenta,      
    urlcolor=cyan,
}

\usepackage{algorithm} 
\usepackage[noend]{algpseudocode}

\newcommand{\R}{\mathbb{R}}

\begin{document}

%

\title{ \fbox{%
  \parbox{1\textwidth}{%
  \small{\textbf{Please cite the journal version instead of this preprint}: \\
Spathis, D., Passalis, N., \& Tefas, A. (2018). Interactive dimensionality reduction
using similarity projections. \textit{Knowledge-Based Systems.}}
\\ \\
\footnotesize{\textcopyright 2018. This manuscript version is made available under the \href{http://creativecommons.org/licenses/by-nc-nd/4.0/}{CC-BY-NC-ND 4.0 license}.
}
}}
\newline \newline \newline Interactive dimensionality reduction using similarity projections}

%
%
%
%

\author[label1,label2]{Dimitris~Spathis \thanks{} }
\cortext[cor1]{Corresponding author. Work done while at the Aristotle University of Thessaloniki.}

\author[label2]{Nikolaos Passalis}

\author[label2]{Anastasios Tefas}

\address[label1]{Department of Computer Science and Technology, University of Cambridge, UK}
\address[label2]{Department of Informatics, Aristotle University of Thessaloniki, Greece}



%
%

\markboth{TVCG SUBMISSION}%
{Shell \MakeLowercase{\textit{et al.}}: Bare Demo of IEEEtran.cls for Computer Society Journals}
%




\begin{abstract}
Recent advances in machine learning allow us to analyze and describe the content of high-dimensional data like text, audio, images or other signals. In order to visualize that data in 2D or 3D, usually Dimensionality Reduction (DR) techniques are employed. Most of these techniques, e.g., PCA or t-SNE, produce static projections without taking into account corrections from humans or other data exploration scenarios. In this work, we propose the \textit{interactive Similarity Projection (iSP)}, a novel interactive DR framework based on similarity embeddings, where we form a differentiable objective based on the user interactions and perform learning using gradient descent, with an end-to-end trainable architecture. Two interaction scenarios are evaluated. First, a common methodology in multidimensional projection is to project a subset of data, arrange them in classes or clusters, and project the rest unseen dataset based on that manipulation, in a kind of semi-supervised interpolation.  We report results that outperform competitive baselines in a wide range of metrics and datasets. Second, we explore the scenario of manipulating some classes, while enriching the optimization with high-dimensional neighbor information.  Apart from improving classification precision and clustering on images and text documents, the new emerging structure of the projection unveils semantic manifolds. For example, on the Head Pose dataset, by just dragging the faces looking far left to the left and those looking far right to the right, all faces are re-arranged on a continuum even on the vertical axis (face up and down). This end-to-end framework can be used for fast, visual semi-supervised learning, manifold exploration, interactive domain adaptation of neural embeddings and transfer learning.\end{abstract}


\maketitle


%

\section{Introduction}\label{sec:introduction}

%
%
%
%
Recent advances in statistical machine learning allow us to tackle hard real-world problems such as machine translation, speech recognition, image captioning and developing self-driving cars \citep{lecun2015deep}. In order to train machine learning models for the aforementioned tasks large-scale data is required. However, the process of data collection and processing is costly since human annotators must validate the \textit{ground truth} of each dataset. This issue is further aggravated by the limited involvement of domain experts, who are also potential users of such systems, in the training process \citep{amershi2014power}.

Interactive systems that learn from users have been proposed during the last decade for several applications, e.g.,  image segmentation \citep{fails2003interactive}. While these systems allowed to manipulate the input or some parameters of the model, they rarely offered ways to interact with the data points \textit{per se}. To further increase the involvement of end-users in training machine learning systems, we also have to think about the interface and cognitive load we provide them with \citep{aggarwal2004human}. In this work we research dimensionality reduction techniques, which provide ways of projecting high-dimensional data in 2D. While dimensionality reduction is used for many purposes, such as to reduce storage space and preprocessing time \citep{liu2012}, we focus on its usage for visualization. By visualizing the data in a two-dimensional (2D) or three-dimensional space (3D), we make it easier for humans to understand the structure of data in a manner that feels natural, exploiting our inherent ability to recognize and understand visual patterns \citep{tongli2001}.

For a dataset that consists of $N$ $n$-dimensional points \mbox{$\mathbf{p}_i \in \mathbb{R}^n$}, dimensionality reduction (DR) can be defined as a function which maps each point ${\textbf{p}}_i \in \mathbb{R}^{n} $ to a low-dimensional point ${\textbf{q}}_i \in \mathbb{R}^{m} $:
\begin{equation}
f_P : \R^{n} \rightarrow \R^{m}
\label{eqn:Dimensionality Reduction}
\end{equation}
Here, $n$ is typically large (from tens to thousands of dimensions), $m$ represents the number of dimensions in the low-dimensional space, typically 2 or 3, while $P$ denotes the parameters of the function used to perform the projection. 

Even though many dimensionality reduction techniques leverage different concepts, most of them share a common property: the objective function used to optimize the final projection is a linear combination of pairwise distances between the data points, merely using only second-order statistics \citep{yan2007graph}. 
However, this renders most of the existing dimensionality reduction techniques prone to outliers, while it increases the difficulty of interacting with the data; it is not always straightforward to manipulate the distance between different points (there is no universally small or large distance). The aforementioned limitations highlight the need for methods that use bounded similarity metrics instead of unbounded distance metrics. An illustrative example is the recent success of t-SNE \cite{maaten2008visualizing}, which transforms distances to probabilities using a non-linearity (Gaussian kernel). That way it effectively addresses the so-called \textit{crowding problem} \cite{maaten2008visualizing}. 
In this work, we propose a novel data visualization and exploration framework that allows visual machine learning and enables layout steerability with a human in the loop.


To the best of our knowledge, the proposed method is the first that uses higher order statistics to model the user interactions, while introducing a structured and meaningful way to steer the projections leading to a versatile interactive visualization framework. This allows overcoming several limitations of existing methods that only rely on second-order statistics and provides a generic framework that can be used for a variety of tasks (as we demonstrate through the paper).
The proposed framework, called \textit{interactive Similarity Projection (iSP)}, is using a recently proposed methodology \citep{passalis} and can be combined with any differentiable projection function, ranging from linear and kernel functions to deep neural networks. The iSP framework is evaluated on six datasets from a diverse range of domains and it is demonstrated that it outperforms other evaluated baselines and state-of-the-art methods. Since accuracy compared to ground-truth is not always the goal, we demonstrate a novel interactive exploratory scenario where we appropriately manipulate a small number of data points to uncover the latent structure of the data, showing the efficiency of our method using both quantitative metrics and qualitative illustrations.

The main contributions of our work are:

\begin{itemize}
\item a proposed end-to-end trainable methodology for layout steerability and semi-supervised classification with a human in the loop

\item the adaptation of modern gradient descent algorithms towards handling interaction scenarios as optimization targets

\item a thorough comparison of available \textit{multidimensional projection} implementations of the relevant literature

\item a novel interaction scenario and methodology based on gradient descent and nearest neighbors that uncover meaningful manifolds by interaction

\end{itemize}

The applications and extensions of the proposed approached are countless. iSP allows to easily derive out-of-sample extensions of competitive projection methods, as well as more complex targets like the output of a convolutional neural network. Another intersting approach would be to interactively customize pretrained word embeddings for tailored tasks (sentiment analysis). Although we are not presenting experiments for all cases we consider them to be easily implemented. We discuss them in Section 5.

The rest of the paper is structured as follows. First, the related work is briefly discussed and compared to iSP in Section~\ref{section:related-work}. Then iSP is presented in detail (Section~\ref{section:proposed-method}) and evaluated (Section 4). Finally, conclusions are drawn and future work is discussed in Section 5.

\section{Related work}
\label{section:related-work}
Our work is built upon widely used dimensionality reduction techniques such as the Principal Components Analysis (PCA) and the \mbox{t-Distributed} Stochastic Neighbor Embedding \mbox{(t-SNE)\citep{maaten2008visualizing}}. To the best of our knowledge the proposed method is the first interactive visualization method that models the user's interactions using higher-order statistics and provides an intuitive and structured way to steer the projections using that information.
That allows for overcoming the limitations of existing methods.

For example, the Dis-Function \citep{brown2012dis} methodology performs gradient descent to maintain the relative distances of points the user did not select while encouraging changes that affect the selected points in the desired direction, but unlike our setting, it operates on euclidean distances. A conceptually similar model, based on Multi-Dimensional Scaling (MDS), is suggested by \citep{endert2011observation}. More generally, this new paradigm of a bi-directional visualization pipeline (display, adjust and repeat) is termed as Visual to Parametric Interaction (V2PI) by \citep{leman2013visual}, in which they adapt common techniques like PCA/MDS, while we operate on a similarity matrix, being agnostic to a specific method. We also note that other visualization methods exist, for the purpose of this paper, we focus on scatter plots with color-coded classes due to user familiarity and performance \citep{bernard2017comparing}. 

A group of literature usually called \textit{multidimensional projection} uses some seeding direct manipulation points (also called \textit{"control points"}), which are a subset of the original data points. The intuition goes that by manipulating a subset of data points, a mapping function is (implicitly or explicitly) learned. Then, this function is used (or approximated) to project the rest of the dataset.
Numerous approaches related to control-point manipulation have been suggested. The \textit{Local Affine Multidimensional Projection} (LAMP) starts by projecting a subset of control points and then interpolates the remaining points through orthogonal affine mappings, using the \textit{Singular Value Decomposition} (SVD) technique \citep{joia2011local}.  The \textit{Part-Linear Multidimensional Projection} (PLMP)~\cite{paulovich2010two} allows for scaling to big datasets by first constructing a linear map of the control points using \textit{Force Scheme}~\citep{tejada2003improved}. Next, this mapping is used to project the remaining points. 
Finally, the \textit{Kernel-based Linear Projection} (KELP) allows for visualizing how \textit{kernel functions} project the data in high-dimensional spaces, while allowing for interacting with the learned projection \citep{barbosa2016visualizing}. 

Most interpolation multi-dimensional techniques like LAMP, that allow to manipulate some labeled instances and project an unlabeled superset thereof, are based on MDS-like solutions combined with eigendecomposition, linear solvers and least squares approaches. We propose to state that problem as an \textit{iterative} interactive optimization problem, in which the user's interaction is used to optimize the projection using gradient descent.

Another substantial area of the literature has been focused in semi - supervised learning. While our methodology can perform semi-supervised learning, there some significant differences. This line of research proposes variants of the same approach: if we have a small training set and a big test set, can the geometry of the latter help the first? For example there have been proposed: extensions of LDA by constructing the nearest neighbor graph for the unlabeled data by cleverly adding a data-dependent regularizer \cite{cai2007semi}; a new discriminative least squares regression \cite{luo2017adaptive};  interesting concepts of soft label propagation based on negation that reminds of the recent success of negative sampling in NLP (\textit{word2vec}) \cite{hou2011semisupervised}, as well as random-walk based label propagation \cite{nie2011semisupervised}.

However, in contrast with our approach they do not take into account any kind of interaction, while producing static projections that cannot be manipulated by the users to expose different views of the data according to their needs. 
In our experiments below (see sect.~\S\ref{section:experiments}), we show that users can iteratively move points around and project the unlabeled dataset until they find the best projection. Our approach does not depends on labels at all, since the user just arranges the data points as he/she wishes and then we appropriately estimate a mapping function that can project the data according to the manipulation provided by the user. This is different from supervised/semi-supervised forms of DR, where labels are required for the training process. Consequently, the semi-supervised capability is just a (fortunate) side-effect of the proposed similarity embedding optimization process (see sect.~\S\ref{section:proposed-method}).

\section{Proposed Method}
\label{section:proposed-method}
Our optimization methodology uses the recently proposed Similarity Embedding Framework (SEF) \cite{passalis}, which we present in this section. Then, we build upon the SEF to derive iSP, a novel fast interactive data visualization technique.

The existing SEF formulation was oriented towards simple non-interactive dimensionality reduction tasks. Therefore, even though SEF provides a powerful DR formulation, directly employing SEF's objective for interactive visualization was not possible. To this end, we proposed new interaction scenarios which allow for fast semi-supervised learning with control points, the possibility to clone other competing interactive DR techniques, as well as a neighbourhood-based method with a novel objective to discover semantic manifolds through iterative interaction. Our mathematical derivations attempt to showcase the optimization process in the simplest possible way, so that the uninitiated reader will understand that we perform gradient descent upon similarity matrices. All in all, our method enables to derive a novel and powerful interactive visualization method that is capable of effectively handling different interaction scenarios that otherwise would not have been possible.

\subsection{Similarity Embedding Framework}
Let $S(\mathbf{x}_i,\mathbf{x}_j)$ be the pairwise similarity between the data points $\mathbf{x}_i \in \mathbb{R}^n$ and $\mathbf{x}_j \in \mathbb{R}^n$. Note that a similarity metric $S$ is a bounded function that ranges between 0 and 1 and expresses the proximity between two points. Then, the similarity matrix of the projected data is defined as $[\textbf{P}]_{i,j} = S(f_{DR}(\mathbf{x}_i), f_{DR}(\mathbf{x}_j))$, where $i$ is a row and $j$ is a column of this matrix and $f_{DR}$ is the projection function.

SEF's main goal is to learn a projection in which the similarities in the low-dimensional space, i.e., in our case the visual space, are as close as possible to a selected ``target''. The target similarity matrix $\textbf{T}$ is a square matrix that can be the result of many methods, such as direct manipulation of data points (as in our case), or other DR techniques (if we want to mimic them), such as PCA, LDA, t-SNE, etc. In order to learn the projection function $f_{DR}(\mathbf{x})$ we optimize the following objective function:
\begin{equation}
J_s = \frac{1}{2 \parallel \textbf{M}\parallel _1}
\sum_{i \neq j}^N [\textbf{M}]_{i,j} ([\textbf{P}]_{i,j} - [\textbf{T}]_{i,j})^2
\label{eqn:Objective function}
\end{equation}
where $\textbf{M}$ is a matrix acting as a weighting mask defining the importance of attaining the target similarity of the data and $\parallel \textbf{M}\parallel_1 = \sum_{i=1}^N  \sum_{j=1}^M  |\textbf{M}_{i,j}| $ is the entrywise $l_1$ norm of matrix $\textbf{M}$. When each data point pair achieves its target similarity, the objective function (\ref{eqn:Objective function}) is minimized, while when a pair has different similarity from its target, it is getting penalized.

Although $\mathbf{T}$ can be any target that we want to achieve during the projection, we use the Gaussian kernel (also known as Heat kernel) to define the similarity between the projected points, i.e., $S(\textbf{x}_i,\textbf{x}_j) = exp(-\parallel \textbf{x}_i - \textbf{x}_j \parallel_2^2 / \sigma_p)$, where $\sigma_p$ acts a scaling factor. Therefore, the similarity matrix $\textbf{P}$ is defined as:
\begin{equation}
[\textbf{P}]_{i,j} = exp(-\parallel f_{DR}(\textbf{x}_i) - f_{DR}(\textbf{x}_j) \parallel_2^2 / \sigma_p)
\label{eqn:Similarity matrix P}
\end{equation}

We can also imitate or \textit{clone} existing DR techniques. Let $c(\textbf{x})$ be a technique to be cloned. Then, we can mimic $c(\textbf{x})$ by setting the target matrix as: 
\begin{equation}
[\textbf{T}]_{i,j} = exp (- \frac{\parallel c(\textbf{x}_i) - c(\textbf{x}_j) \parallel_2^2 }{\sigma_{copy}})
\label{eqn:Clone_target}
\end{equation}
where $\sigma_{copy}$ is the scaling factor used to calculate the similarities between the low-dimensional points, as projected using the techniques that is to be cloned.

The projection function $f_{DR}$ could be defined in multiple ways, ranging from simple linear transformations to non-linear methods such as kernel projections and deep neural networks. In order to minimize the loss in objective function~(\ref{eqn:Objective function}), gradient descent is used. Therefore, it is required to calculate the derivative of the objective function $J_s$ with respect to the parameters of each projection.

One of the simplest projection methods is the linear transformation of the input space, such that $f_{DR}(\textbf{x}) = \textbf{W}^T \textbf{x}$, where $\textbf{W} \in \mathbb{R}^{n \times m}$ is the projection matrix. Let the relationship between the original data $\textbf{X}$ and the projected $\textbf{Y}$ be that of $\textbf{y}_i = f_{DR}(\mathbf{x}_i)$. 
The derivative of the objective function $J_s$ when a linear transformation is used is calculated as:
\begin{equation}
\label{eq:derivative}
\frac{\partial J_s}{\partial [\textbf{W}]_{k,t}} 
= 
\frac{1}{\parallel \textbf{M}\parallel _1} \sum_{i=1}^N  \sum_{j=1}^N [\textbf{M}]_{i,j} ([\textbf{P}]_{i,j} - [\textbf{T}]_{i,j}) \frac{\partial [\textbf{P}]_{i,j}}{\partial [\textbf{W}]_{k,t}} 
\end{equation}
where 
\begin{equation}
\frac{\partial [\textbf{P}]_{i,j}}{\partial [\textbf{W}]_{k,t}} = -\frac{2}{\sigma_p} [\textbf{P}]_{i,j} ([\textbf{Y}]_{i,t} - [\textbf{Y}]_{j,t}) ([\textbf{X}]_{i,k} - [\textbf{X}]_{j,k})
\end{equation}
The objective function~(\ref{eqn:Objective function}) is optimized using gradient descent: 
\begin{equation}
\Delta \textbf{W}  = - \eta \frac{\partial J}{\partial \textbf{W}} 
\label{eqn:gradient descent}
\end{equation} where $\eta$ is the learning rate. In this work the Adam algorithm is used for the optimization~\citep{kingma2014adam}, since it has been proved to be fast and reliable. The gradient descent learning rate of Adam follows the current standards in deep learning training (0.001).

On the other hand, kernel methods are known to provide superior solutions, since they transform the input space into a higher dimensional one in order to solve the problem in a linear manner there. Let $\boldsymbol{\Phi} = \phi(\textbf{X})$ be the matrix of data in the high dimensional space (also known as Hilbert space), where $ \phi(\mathbf{x})$ is a function that projects the data in a higher dimensional space. In a similar way with the linear version, we seek to learn a linear mapping from the Hilbert space into the visual space. 
Therefore, the matrix $\mathbf{W}$ is redefined as:
\begin{equation}
\label{eq:weights-kernel}
\mathbf{W} = \phi(\mathbf{X})^T \mathbf{A} = \mathbf{\Phi}^T \mathbf{A}
\end{equation}
where $\mathbf{A} \in \mathbb{R}^{n \times m}$ is a coefficient matrix that defines each projection as a linear combination of the data points. The projection can be now calculated, using the Eq. (\ref{eq:weights-kernel}) as: 

\begin{equation*}
\mathbf{Y}^T = \mathbf{W}^T \mathbf{\Phi}^T = \mathbf{A}^T \mathbf{\Phi} \mathbf{\Phi}^T = \mathbf{A}^T\mathbf{K}
\end{equation*}
where $\mathbf{K} = \mathbf{\Phi} \mathbf{\Phi}^T  \in \mathbb{R}^{n \times n}$ is the kernel matrix of the data that contains the inner products between the data points in the Hilbert space, i.e., $[\mathbf{K}]_{ij} = \phi(\mathbf{x}_i)^T\phi(\mathbf{x}_j)$.  When a Reproducing Kernel Hilbert Space (RKHS) is used the kernel matrix can be calculated without explicitly calculating the inner products in the Hilbert space. Therefore, different choices for the kernel matrix $\mathbf{K}$ lead to different Hilbert spaces. Among the most used kernels is the Gaussian/RBF kernel, i.e., $[\mathbf{K}]_{ij} = exp(-||\mathbf{x}_i-\mathbf{x}_j||_2^2 / \gamma^2)$, which maps the input points to an infinite dimensional space, and it is the kernel that was used in our experiments. The kernel width (or sigma) is the mean of the pairwise distances of the data. Since this parameter depends on each dataset, it was not tuned or changed during our experiments and therefore cannot influence the results. Several other kernels functions have been proposed in the literature, e.g., polynomial kernels \cite{smola1998learning} or a combination thereof \citep{liu2015}, and they can be also used with the iSP method.

Therefore, in the kernelized method the coefficient matrix $\mathbf{A}$ is to be learned, instead of the weight matrix $\mathbf{W}$. Again, the gradient descent algorithm is used and the corresponding gradient is derived as:

\begin{equation}
\label{eq:ksp-gradient}
\frac{\partial J}{\partial [\mathbf{A}]_{k,t}} = \frac{1}{||\mathbf{M}||_1}  \sum_{i=1}^N \sum_{j=1}^N [\mathbf{M}]_{i,j}([\mathbf{P}]_{i,j}-[\mathbf{T}]_{i,j}) \frac{\partial [\mathbf{P}]_{i,j}}{\partial [\mathbf{A}]_{k,t}}  
\end{equation}
where 
\begin{equation*}
\frac{\partial [\mathbf{P}]_{i,j}}{\partial [\mathbf{A}]_{k,t}}   = -\frac{2}{\sigma_P}[\mathbf{P}]_{i,j} ([\mathbf{Y}]_{i,t} - [\mathbf{Y}]_{j,t} ) ([\mathbf{K}]_{i,k} - [\mathbf{K}]_{j,k})
\end{equation*}

\subsection{Interactive Similarity Projection}

\begin{figure}[H] 
	\centering
	\includegraphics[width=1.0\textwidth]{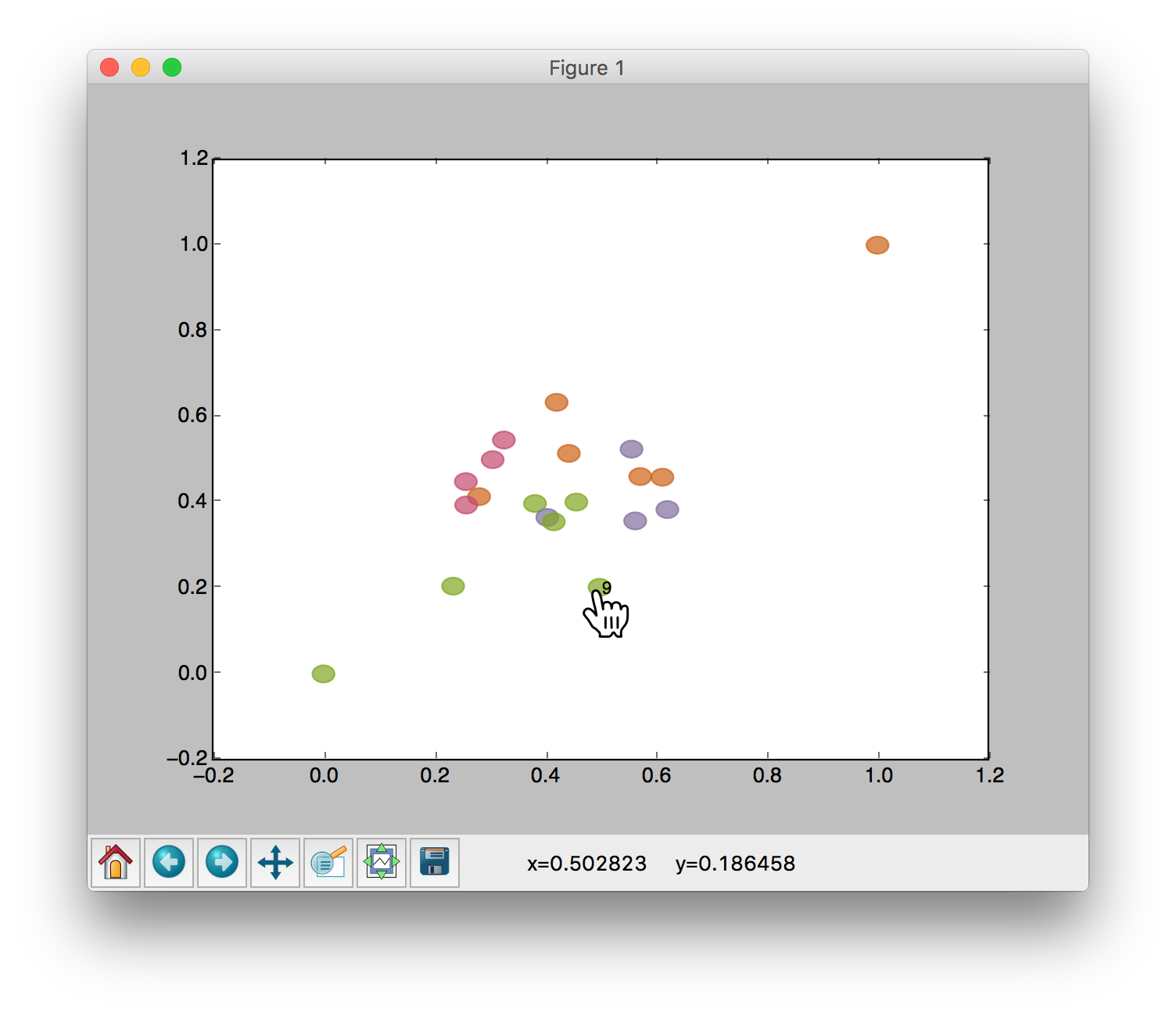}
	\caption{Screen-shot of the interactive scatter plot built to enable users to manipulate data points. Along with the color coded classes information (in case of labeled data), the user can see the label of the data point currently being dragged (in that case the digit 9).}
	\label{fig:scatterplot}
\end{figure}

Here we present the two interaction methodologies that we tested on a custom scatter plot. Considering the real-world practitioners and data scientists today, the prototype interactive scatter plot (Fig.~\ref{fig:scatterplot}) that complements our study was built in Matplotlib \citep{hunter2007matplotlib}. The supported interactions are dragging individual data points and inspecting their label.

The two different methods that are presented are used for two distinct concepts. We noticed that interaction was used in the literature for semi-supervised learning so that the user deals with a couple of data points at a time. Also, another concept was the data exploration after projecting the initial dataset in 2D. There are not advantages of the one over the other since they could act complimentary; data scientists would explore data along with domain experts and then perform classification/clustering on an interactive manner. Therefore we propose two methods that are based in gradient descent through interaction.

\subsubsection{Interaction scenario \#1: Semi-supervised learning
based on control points}
The proposed interactive dimensionality reduction learning algorithm is summarized in Algorithm~\ref{alg:iSIL_algorithm}. More formally, let $\textbf{X} = [ {\textbf{x}_i,...,\textbf{x}_N}]$ be the data matrix (in the original feature space) and $\textbf{X}_s = [{\textbf{x}_{s_i},...,\textbf{x}_{s_{N_i}}}]$ a subset of the data that is used as control points. Let $\textbf{Y}_s$ be the projection of control points $\textbf{X}_s$ in the visual space and $\textbf{Y}$ the final projection of $\textbf{X}$ in the visual space. First, the projection is initialized by cloning an existing technique (lines 2-3). The user interacts and manipulates the control points $\textbf{Y}_s$ and their respective coordinates producing $\widetilde{\textbf{Y}_s}$ (lines 4-5). Finally, the previous projection is optimized according to the user's interaction (lines 6-7) and the whole dataset is visualized by calculating  $\widetilde{\textbf{Y}}$ (line 8). Note that tilde'd  characters are used to denote the results after the manipulation. 

 Since our technique is iterative, the user can correct some samples, observe the result and continue correcting the respective outcomes. To not overwhelm users, we display a subset of the dataset. This also makes sense in cases where a small subset might be labeled. 

In any case, with our method the user has to wait for the pre-defined number of steps for the stochastic gradient descent to be completed. Then they can observe the result and re-adjust. Since the algorithm may converge earlier, we can use some kind of early stopping \citep{prechelt1998automatic} which is getting popular with neural networks lately. However, there is still open research on how to integrate intermediate results from algorithm iterations, than just computing the objective in every step with the new positions of the data points \citep{kim2017pive}.

\begin{algorithm}
\caption{Interactive Similarity-based Interpolation Learning Algorithm}\label{alg:iSIL_algorithm}
\begin{algorithmic}[1]
\Require {A matrix $\textbf{X} = [\textbf{x}_i,...,\textbf{x}_N]$ of $N$ data points, the subset of the control points $\textbf{X}_s$, and a technique $s(\mathbf{x})$ that can be used for initializing the projection}
\Ensure {Projected dataset $\widetilde{\textbf{Y}}$.}
\Procedure{InteractiveDRLearning}{}
\State $\boldsymbol{T} \gets$ clone (s($\mathbf{X}_s$)) \Comment{Clone Force Scheme, t-SNE, or any other DR technique using Eq. \ref{eqn:Clone_target}.}
\State $\boldsymbol{W} or \boldsymbol{A} \gets$ SimilarityEmbeddingLearning($\boldsymbol{X}_s, \boldsymbol{T}$) \Comment{Learn projection using Eq.~\ref{eq:derivative}}

\State $\boldsymbol{Y_s} \gets$ project ($\boldsymbol{X}_s$) \Comment{Project control points using projection matrix \textbf{W} or \textbf{A}.}

\State $\widetilde{\boldsymbol{Y}_s} \gets$ manipulate ($\boldsymbol{Y}_s$) \Comment{Adjust control points on interactive scatter plot.}

\State $\widetilde{\boldsymbol{T}} \gets$  clone ( $\widetilde{\boldsymbol{Y_s}}$) \Comment{Clone control points using again Eq. \ref{eqn:Clone_target}.}

\State $\widetilde{\boldsymbol{W}} or \widetilde{\boldsymbol{A}}  \gets$ SimilarityEmbeddingLearning($\widetilde{Y_s}, \widetilde{T}$) \Comment{Learn projection using Eq. 6.}

\State $\widetilde{\boldsymbol{Y}} \gets$ project ($\boldsymbol{X}$) \Comment{Project original dataset using projection matrix $\widetilde{\boldsymbol{W}}$ or $\widetilde{\boldsymbol{A}}$}.

\EndProcedure
\State \textbf{return} the projected dataset $\widetilde{\boldsymbol{Y}}$

\end{algorithmic}
\end{algorithm}

\subsubsection{Interaction scenario \#2: Unveiling semantic manifolds through high-dimensional neighbors}

By using the iSP, instead of just cloning the target similarity of the user manipulation, we can learn more from it. Humans tend to spot outliers when comparing similar plots, where these outliers are mostly instances projected inside other classes, far from their respective class, or both cases simultaneously.

Drawing inspiration from the Checkviz framework \citep{lespinats2011checkviz}, we can repurpose iSP to take into consideration these subtle aspects. Checkviz suggested that the two major mapping distortions occur as \textit{false neighborhoods} and \textit{tears}. False neighborhoods appear when a large distance between instances in the high-dimensional space becomes a small distance in the visual space. Subsequently, a tear occurs when a small distance in the original space becomes a large
distance in the projection space (true neighbors are
mapped far apart). 

A naive way to solve the outliers problem described above, would be to take into account the label of each sample, so that by moving an outlier we would bring it even closer to its class, similar to LDA discussed earlier. But that implies that we have class labels for every dataset in our disposal. In a real-world scenario, most datasets are unlabeled or semi-labeled. Also, users frequently move some data points with respect to some other “unmoved” data points that they consider as spatially contextual. However, as \citep{hu2013semantics} points out, these points are not explicitly identified
when directly manipulating the moved points. Therefore, we propose to use the under-exploited information of the high-dimensional neighbors of a given sample we interact with. 

The proposed algorithm is coined as \textit{interactive similarity-based neighbor embedding learning}. More formally, the data $\textbf{X}$ are loaded. The target matrix $\textbf{T}$ is initialized by cloning another technique such as t-SNE or PCA. Then, we perform gradient descent with $\textbf{X}$ and the respective target matrix $\textbf{T}$ in order to minimize the objective function, so that each pair of the projected points achieves its target similarity.

\begin{algorithm}
\caption{Interactive Similarity-based Neighbor Learning Algorithm}\label{alg:iSNEL_algorithm}
\begin{algorithmic}[1]
\Require {A set $\textbf{X} = \{\textbf{x}_i,...,\textbf{x}_N\}$ of $N$ data points.}
\Ensure {Projected points $\widetilde{\boldsymbol{Y}}_{knn}$.}
\Procedure{InteractiveNeighborLearning}{}
\State $\boldsymbol{T} \gets$ clone (technique) \Comment{Clone a DR technique using Eq. \ref{eqn:Clone_target}.}
\State $\boldsymbol{W} or \boldsymbol{A} \gets$ SimilarityEmbeddingLearning($\boldsymbol{X}, \boldsymbol{T}$) \Comment{Learn projection using Eq. 6.}

\State $\boldsymbol{Y} \gets$ project ($\boldsymbol{X}$) \Comment{Project points based on projection matrix $\boldsymbol{W} or \boldsymbol{A}$.}

\State $\widetilde{\boldsymbol{Y}} \gets$ manipulate ($\boldsymbol{Y}$) \Comment{Adjust points on interactive scatter plot.}

\State $ind \gets$  $\widetilde{\boldsymbol{Y}} \setminus \boldsymbol{Y}$ \Comment{Keep an index of the position of  manipulated points.}

\State $\boldsymbol{N}_x \gets$ neighbors ($[\boldsymbol{X}]_{ind}$) ,  $\boldsymbol{N}_y \gets$ neighbors ($\widetilde{[\boldsymbol{Y}]}_{ind}$) \Comment{Find original and visual neighbors for every manipulated point.}

\State $\widetilde{[\boldsymbol{T}]}_{i,j} = 1 : j \in \textbf{N}_x, \widetilde{[\boldsymbol{T}]}_{i,j} = 1 : j \in \textbf{N}_y$


\State $\widetilde{[\boldsymbol{M}]}_{i,j} = 1 : j \in \textbf{N}_x, \widetilde{[\boldsymbol{M}]}_{i,j} = 0.5 : j \in \textbf{N}_y$


\State $\widetilde{[\boldsymbol{M}}]_{ind} \gets$ 1  \Comment{Set weighting mask to 1 for all manipulated points.}

\State $\widetilde{\textbf{W}} or \widetilde{\textbf{A}}  \gets$ SimilarityEmbeddingLearning($\textbf{X}, \widetilde{\textbf{T}}, \widetilde{\textbf{M}}$) \Comment{Calculate Eq. 6 with new target and mask.}

\State $\widetilde{\textbf{Y}}_{knn} \gets$ project ($\textbf{X}$) \Comment{Project original dataset based on new projection matrix $\widetilde{\textbf{W}}$ or $\widetilde{\textbf{A}}$}.

\EndProcedure
\State \textbf{return} projected points $\widetilde{\textbf{Y}}_{knn}$

\end{algorithmic}
\end{algorithm}

After the gradient descent, original data points $\textbf{X}$ are transformed to $\textbf{Y}$ and projected to the visual space. The projected points $\textbf{Y}$ are manipulated and steered by the user on the interactive scatter plot. The new points are denoted as $\widetilde{\textbf{Y}}$ (lines 5-6).

The next step is crucial for the proposed methodology. The indices of the manipulated instances are stored. Original neighbors $\textbf{N}_x$ of the manipulated instances as well as neighbors in the projected space $\textbf{N}_y$ are estimated. The target similarity matrix $\widetilde{\textbf{T}}$ is adjusted so that $\widetilde{[\textbf{T}]}_{i,j}$ = 1 for visual neighbors $\textbf{N}_y$, with the intuition that when we move a data point towards other points, we consider it similar to them. Also the mask $\textbf{M}$ is set to a low value (0.5), as a regularizer (lines 6-10). Moreover, $\widetilde{[\textbf{T}]}_{i,j}$ = 1 for original neighbors $\textbf{N}_x$. The intuition is that the original neighbors will be dragged near the manipulated points. Even if original neighbors do not belong to the same class, they might gravitate towards the manipulated point, eventually revealing a new manifold or a new dimension. Also, in order to increase the importance of the manipulated points, the weighting mask $\textbf{M}$ is set to 1 for them.

The new similarity matrix $\widetilde{\textbf{T}}$ is getting optimized with gradient descent in order to find a new projection matrix. The intuition is that the learning algorithm will iteratively learn the layout suggested by the user manipulation and the neighbors, revealing interesting attributes of the data, while \textit{spreading} that optimization to the rest of dataset that does not belong to the neighbors of the manipulated points.

Conceptually, this technique tries to combine local and global neighborhood preservation. LoCH \citep{fadel2015loch} attempted something similar by first projecting the control points which are clustering centers and then performing an iterative approximation in order to place
each data point close to the \textit{convex hull} of its nearest neighbors seeking to maintain
small neighborhood structures. However, LoCH does not take into account interactive scenarios but focuses on where to place the neighboring points on the 2D space.

\section{Experiments}
\label{section:experiments}
The used methodology, the implementation details and some interaction scenarios were described in the previous Section. In this Section, a thorough investigation of the iSP is presented. Evaluation is conducted with representative datasets, covering a wide range of sizes, dimensions and domains. 

\subsection{Datasets}
For evaluating the iSP, the following datasets from a wide range of domains are used: a multi class \textit{handwritten digit} recognition dataset, a multi class \textit{text classification} dataset, a multi class \textit{head pose} estimation dataset, a multi class \textit{outdoor image segmentation} dataset, a two class \textit{breast cancer diagnosis} dataset and a three class \textit{chemical--wine} recognition dataset. They were selected due to their variety of classes (from 2 to 20), size (from 178 to 60000), and dimensions (from 13 to 12000). These  datasets are quite common in the dimensionality reduction and machine learning literature, allowing us to compare our findings with the research community. Other visualization techniques exist for different kinds of data, such as time-series \citep{li2013,Ltifi2016}.

Please note that as a pre-processing step we perform PCA to \textit{whiten} the data and reduce the collinearity of features. It's a standard technique before applying even state-of-art methods like tSNE since it produces more stable projections. Quoting the tSNE \citep{maaten2008visualizing} paper:\textit{ In all of our experiments, we start by using PCA to reduce the dimensionality of the data to 30. This speeds up the computation of pairwise distances between the datapoints and suppresses some noise without severely distorting the interpoint distances. We then use each of the dimensionality reduction techniques to convert the 30-dimensional representation to a two-dimensional map and we show the resulting map as a scatterplot}. Because we compare different datasets, by keeping the number of features that explain 90\% of the variance allows us to make sure that all compared algorithms take as input significant features only.

The \textbf{MNIST}\footnote{\url{http://yann.lecun.com/exdb/mnist}} dataset of handwritten digits contains 60000 images for training and 10000 for testing \citep{lecun1998mnist}, while covering 10 classes (0 to 9). MNIST is a natural candidate for dimensionality reduction, since the digit images are high-dimensional ($28 \times 28 = 784$ dimensions), but much of the variation of data can be explained by a smaller number of \textit{semantic} features such as height, tilt, orientation and roundness. We use the raw pixel representation and reduce the dimensionality with PCA, by keeping 90\% of the original variance. Experiments were performed with a subset of classes (digits \textit{2, 4, 7, 9}).

Another format that is represented in multidimensional --usually sparse-- data, is text. One of the standard benchmark datasets for evaluation of text classification and clustering algorithms is the \textbf{20Newsgroups}\footnote{\url{http://scikit-learn.org/stable/modules/generated/sklearn.datasets.fetch_20newsgroups.html}}. This is a collection of 18846 newsgroup documents (Usenet discussion groups), partitioned almost evenly across 20 different topics. These topics cover a range of discussion areas, from computing, to politics, sports, and science. For the experiments, a subset of documents and categories is chosen. We keep 500 documents belonging to classes 1 (alt.atheism), 6 (comp.windows.x), 10 (rec sport baseball), 15 (sci.space), and 20 (talk.religion.misc). The reasoning behind that choice is that these categories cover almost every topic of the 20 categories, while 2 of them (atheism and religion) might be semantically related. For the vector representation of text, many options were evaluated, like word frequencies, log-scaled word frequencies and n-grams. However, since these methods do not capture high level semantics of text, we ran the documents through a pre-trained Google News word2vec model\footnote{\url{https://code.google.com/archive/p/word2vec/}}, obtaining a 300-dimensional vector for each word \citep{mikolov2013distributed, rehurek_lrec}. Then we averaged the vectors for each document.

In many computer vision tasks, we need to know how the head is tilted or oriented with respect to a camera. A standard dataset to evaluate head pose estimation is the \textbf{Head Pose Image}\footnote{\url{http://www-prima.inrialpes.fr/perso/Gourier/Faces/HPDatabase.html}} Database \citep{gourier2004estimating}, a benchmark of 2790 monocular face images of 15 people with variations of pan and tilt angles from -90 to +90 degrees. For every person, 2 series of 93 images (93 different poses) are available. As a pre-processing step, the face areas of each person were isolated, following the coordinates provided by the dataset metadata. The resulting image is of size 64x64 pixels, covering 3 RGB channels. As a result the final dimensionality of each image is 12288 (64x64x3). The raw pixel representation is used as feature vector while the initial dimensions were reduced with PCA, so that the explained variance is 90\%. 

Image segmentation is an ongoing challenge in computer vision research, with the objective being the partitioning of an image into multiple segments \citep{pal1993review}. The \textbf{Image Segmentation} \footnote{\url{https://archive.ics.uci.edu/ml/datasets/Image+Segmentation}} dataset contains 2100 samples with 19 dimensions, and 300 samples per class.  The samples were drawn randomly from a database of 7 outdoor images, which were hand-segmented to create a classification for every pixel. Each instance is a 3x3 region. These 7 outdoor images correspond to the 7 classes of \textit{brickface, sky, foliage, cement, window, path, grass}.

Breast cancer is a disease that develops from breast tissue. Signs of breast cancer may include lumps on the breast or a change in breast shape. Therefore, imaging data can be used to diagnose it early. One of the most common datasets on that regard is the \textbf{Breast Cancer Wisconsin}\footnote{\url{https://archive.ics.uci.edu/ml/datasets/Breast+Cancer+Wisconsin+(Diagnostic)}} (Diagnostic) dataset \citep{street1993nuclear}. It contains 569 images and 30 features, with class distribution of 357 benign and 212 malignant instances.

Chemists test different characteristics of wine in order to evaluate its quality. The \textbf{Wine}\footnote{\url{https://archive.ics.uci.edu/ml/datasets/Wine}} dataset is the result of a chemical analysis of wines grown in the same region in Italy, derived from three different cultivars. The analysis determined the quantities of 13 constituents found in each of the three types of wines. The instances are 178 and the class distribution is class 1: 59, class 2: 71, class 3: 48.

\begin{table}
\caption{Description of evaluation datasets.}
\label{tab:datasets}
\centering
\begin{tabular}{l l l}
\toprule
Dataset & Size & Dimensions\\
\midrule
MNIST & 60000 & 784\\
Newsgroup & 18282 & 300\\
Head Pose & 2790 & 12288\\
Segmentation & 2100 & 19\\
Cancer & 569 & 30\\
Wine & 178 & 13\\
\bottomrule\\
\end{tabular}
\end{table}

\subsection{Evaluation protocol}
Following the methodology presented on the previous section, experiments with the datasets described in the previous Subsection (summarized in Table \ref{tab:datasets}) were performed in order to evaluate the iSP against other baselines and state-of-the-art methods. Two interaction scenarios are examined: semi-supervised interpolation based on control points, where we move every data point towards its class center (\S 3.2.1) and exploration based on original neighbors, where we move a class or classes far from the rest points (\S 3.2.2). In order to compare the first scenario with others, the manipulated control points $\widetilde{Y_s}$, data $X$ and control points in original dimensions $X_s$ are fed into LAMP~\cite{joia2011local}, PLMP~\cite{paulovich2010two}, and KELP~\cite{barbosa2016visualizing}. These specific methods are picked due to their recency, similarity with the proposed technique and --last but not least-- availability of their implementations in public code repositories.
 
In the conducted experiments we estimate the class separation, clustering assignment, and neighbor error. The class separation is evaluated with the {nearest centroid} algorithm. In the \textbf{Nearest Centroid} algorithm, each class is represented by one centroid and test samples are assigned to the class with the nearest centroid \citep{tibshirani2002diagnosis}. For multi-class problems, we follow the "weighted average" method in order to calculate the mean classification precision. Clustering assignment is measured with the  {\textbf{Silhouette coefficient}} \cite{rousseeuw1987silhouettes}, that measures the cohesion and separation between grouped data. In order to visualize the {\textbf{average neighbor error}} per data point, we calculate the 10 neighbors of each point in the visual space and we sum their euclidean distance in the high-dimensional space. This score is then normalized to [0,1], so that a point that its 2D neighbors are far in the feature space, gets a high error close to 1. 

\subsubsection{Interaction scenario \#1: Evaluating semi-supervised classification, clustering and neighbor position}
To evaluate this scenario, we test Wine, Cancer and Segmentation datasets with their original sizes and dimensions, while a subset of 1240 digits of MNIST is chosen. These digits belong in four classes: \textit{"2", "4", "7", "9"}. The number of control points is the  $\sqrt[]{n}$ of their original dataset sizes, as proposed in the literature \cite{paulovich2010two}, so we manipulate 14 random control points on Wine dataset, 24 on Cancer, 46 on Segmentation and 35 on MNIST. Choosing the best control points for visualization purposes is beyond the scope of this paper and it is an interesting future work combining with active learning for example. We should also note that we run the experiments 10 times and we report average metrics and standard deviation. 

The experimental procedure starts with an initialization algorithm that projects the control points (PCA, tSNE and Force Scheme), then we manipulate the control points towards their class centers, and finally the whole dataset is projected, according to the mapping found by each technique. Please note that the manipulation of the control points is simulated to allow for a fair evaluation and faster experimentation. We plan to use real users for this task as future work. Below are the results of Nearest Centroid precision on interpolation.

\begin{table} 
\centering
\caption[Nearest centroid results]{Mean classification precision of interpolated datasets after manipulation of control points. Best results marked bold. Standard deviation of 10 runs in parenthesis.}
\label{ncentroid}
\resizebox{1.0\textwidth}{!}{\begin{tabular}{ccccccc} \toprule
\multirow{2}{*}{Data} & \multirow{2}{*}{Init} & \multicolumn{4}{c}{Nearest Centroid} \\
 &  & \multirow{2}{*}{KELP} & \multirow{2}{*}{LAMP} & \multirow{2}{*}{PLMP} & \multicolumn{2}{c}{Proposed} \\
 &  &  &  &  & Linear & Kernel \\ \midrule  \multirow{3}{*}{Wine} & PCA & 66.11 (6.71) &  59.86 (10.42) &  58.79 (9.93) &  \textbf{81.98} (7.23) &  71.56 (4.09) \\
 & tSNE & 61.42 (4.95) & 63.52 (7.23) & 52.63 (8.92) & 62.50 (13.04) & \textbf{70.19 (4.44)} \\
 & Force & 62.76 (11.07)  & 61.10 (11.45)  & 52.45 (9.30) & 57.97 (7.03)  & \textbf{63.95 (6.33) } \\ \midrule
 \multirow{3}{*}{Cncr} & PCA & 89.84 (2.21) & 83.38 (3.22) & 63.04 (6.12) & 71.77 (7.83) & \textbf{90.68 (0.75)} \\
 & tSNE & 87.91 (5.01) & 82.87 (2.54) & 61.74 (5.35) & 72.17 (4.94) & \textbf{90.95 (1.80)} \\
 & Force & 86.99 (5.31) & 83.97 (1.78) & 59.91 (4.89) & 74.37 (5.34) & \textbf{87.84 (8.14)} \\ \midrule
 
 \multirow{3}{*}{Segm.} & PCA & 58.37 (6.01) & 50.32 (6.33) & 44.20 (8.84) & 50.45 (11.05) & \textbf{60.98 (5.52)} \\
 & tSNE & 62.43 (4.79) & 53.91 (4.11) & 44.24 (6.96) & 55.59 (11.72) & \textbf{64.77 (3.50)} \\
 & Force & 59.70 (5.60) & 53.44 (4.95) & 39.43 (8.79) & 54.71 (6.06) & \textbf{63.40 (4.22)}   \\ \midrule
 
\multirow{3}{*}{MNIST} & PCA & 63.12 (5.59) & 72.79 (5.84) & 29.67 (4.30) & 36.04 (3.10) & \textbf{77.92 (4.53)} \\
 & tSNE & 59.85 (11.84) & \textbf{74.00 (3.43)} & 29.72 (3.62) & 39.11 (4.06) &68.57 (5.73)\\
 & Force & 60.87 (10.52) & 72.20 (5.32) & 31.76 (2.31) & 45.40 (6.32) & \textbf{75.52 (5.90)} 
\\ \bottomrule
\end{tabular}}
\end{table}

The class separation evaluation results are shown in Table~\ref{ncentroid}. The proposed technique, especially when used with a kernel-based projection function, almost always outperforms all the other compared techniques, regardless the used initialization scheme and dataset.  
\begin{figure} 
\centering
\includegraphics[width=1.0\textwidth]{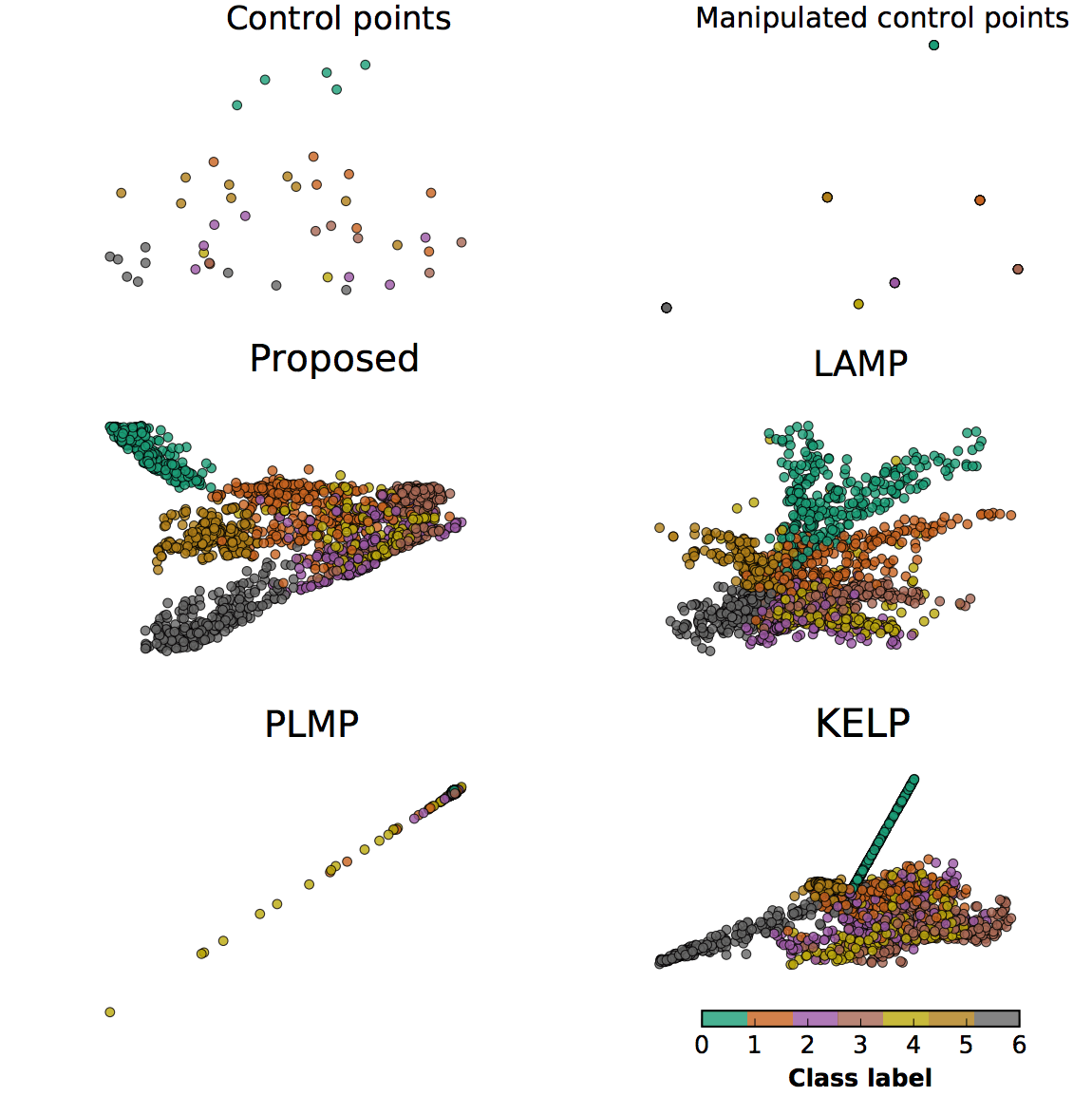}
\caption[Nearest Centroid Plot -- Segmentation Dataset]{Interpolation experiment on the Image Segmentation dataset. Classes correspond to the following labels: green -- sky, orange -- cement, purple -- window, bordeaux -- brickface, yellow -- foliage, mustard -- path, gray -- grass. }
\label{fig:segmentation}
\end{figure}

\begin{figure}
\centering
\begin{minipage}{.5\textwidth}
  \centering
  \includegraphics[width=1.0\linewidth]{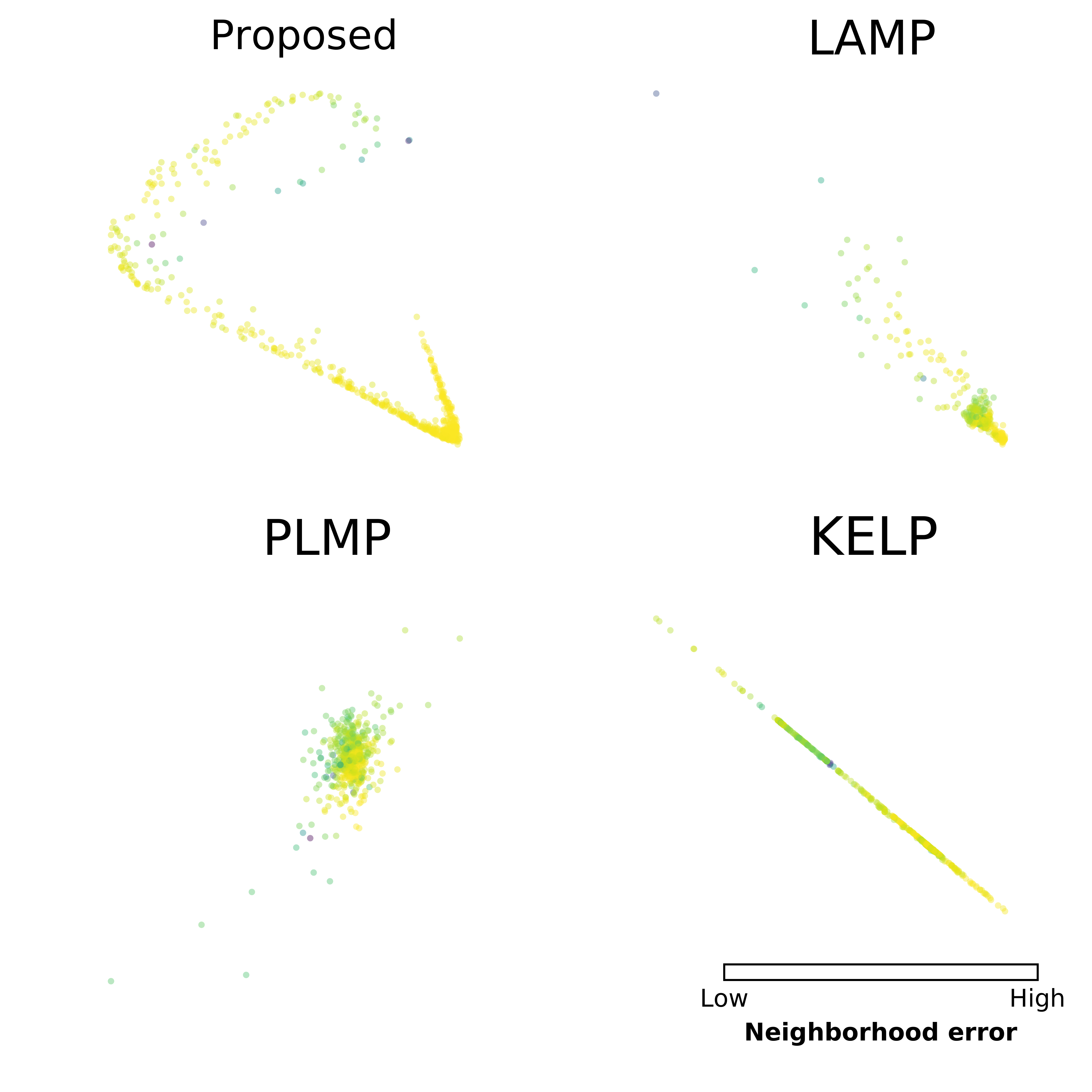}
	\caption{Cancer dataset (neighbor error)}
	\label{fig:cancer-err}
\end{minipage}%
\begin{minipage}{.5\textwidth}
  \centering
  \includegraphics[width=1.0\linewidth]{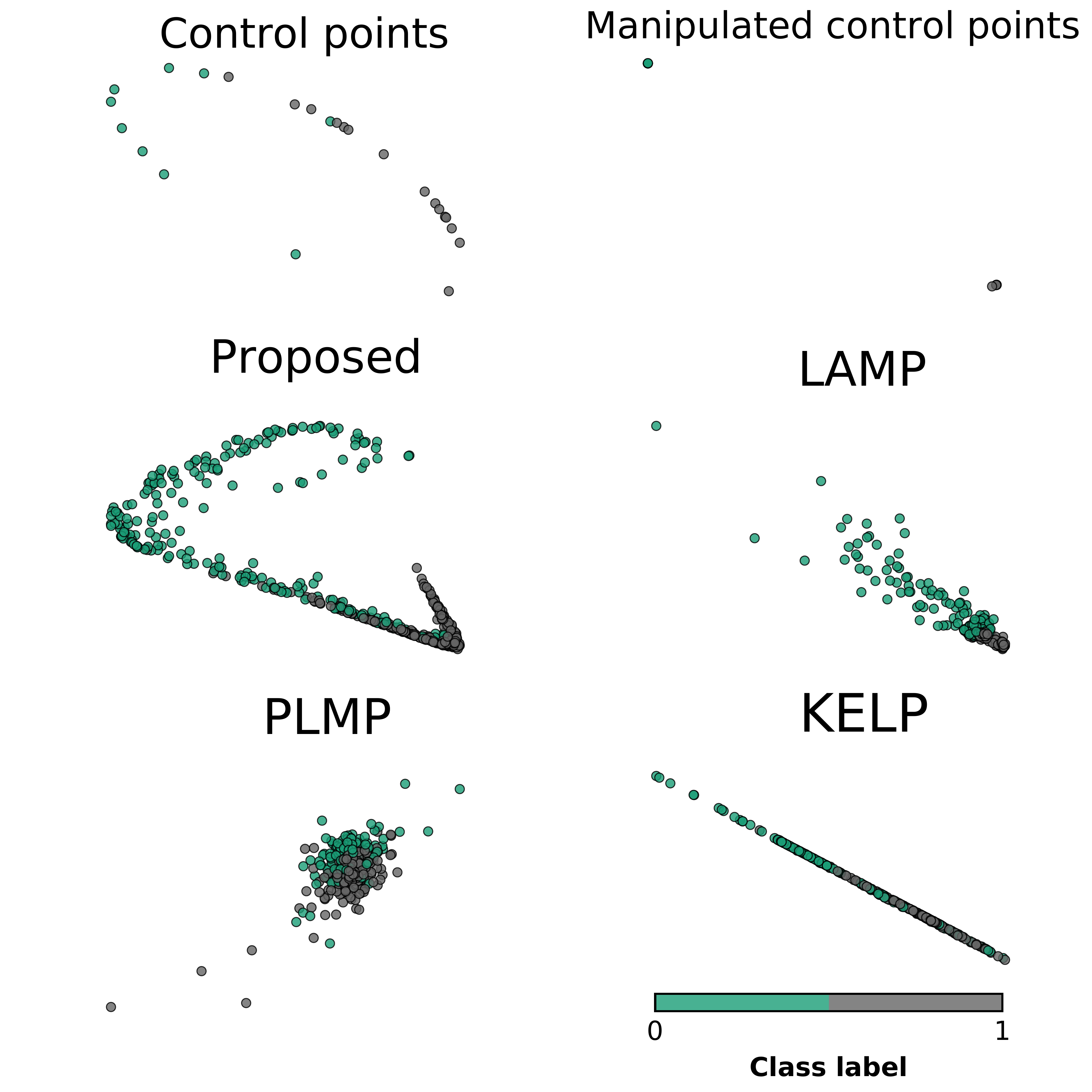}
\caption{Cancer dataset (class separation)}
	\label{fig:cancer}
\end{minipage}
\end{figure}

Fig.~\ref{fig:segmentation} illustrates the projected points along with their classes for all the evaluated techniques using the Segmentation dataset (the kernel variant is used for the proposed method), while in Fig.~\ref{fig:cancer} the Cancer dataset is visualized (again the kernel variant of the proposed method is used). The proposed method always leads to better class separation compared to the proposed methods, as well as it provides more stable behavior on previously unseen data (note that the PLMP and the KELP methods collapses in some cases).

Next, we evaluate the proposed method using a clustering setup (the number of clusters equals the number of classes) using the silhouette coefficient which attempts to count how tightly grouped all the data in each cluster are. Note that the silhouette coefficient normally ranges from -1 to 1. However, to improve the readability of results, values have been scaled to [-100, 100].  The results are reported in Table~\ref{silhouete}. The proposed technique achieves the highest score in every dataset, while the kernel version of the proposed technique still greatly outperforms the linear variant.

\begin{table}[H] 
	\centering
	\caption[Silhouette Coefficient results]{Average silhouette coefficient after manipulation of control points. Best results marked bold. Standard deviation of 10 runs in parenthesis.}
	\label{silhouete}
	\resizebox{1.0\textwidth}{!}{\begin{tabular}{ccccccc} \toprule
			\multirow{2}{*}{Data} & \multirow{2}{*}{Init} & \multicolumn{4}{c}{Silhouette Coefficient} \\
			&  & \multirow{2}{*}{KELP} & \multirow{2}{*}{LAMP} & \multirow{2}{*}{PLMP} & \multicolumn{2}{c}{Proposed} \\
			&  &  &  &  & Linear & Kernel \\ \midrule \multirow{3}{*}{Wine} & PCA & 12.52 (7.00) &  -9.28 (3.69)&  0.55 (4.63) & \textbf{ 28.13 (12.22)} &  19.18 (4.23)\\
			& tSNE & 11.95 ( 6.58) & -14.70 (4.56) & -0.76 (6.05) & 8.23 (8.58) & \textbf{14.49 (8.72)} \\
			& Force & \textbf{15.96 (7.47)}  & -11.41 (5.29)  & 1.60 (4.44) & 4.71 (4.75)  & 12.11 (9.35)  \\ \midrule
			\multirow{3}{*}{Cncr} & PCA & 55.15 (13.60) & 34.14 (4.26) & 12.22 (5.49) & 19.26 (7.63) & \textbf{57.93 (4.75)} \\
			& tSNE & 53.75 (13.78) & 36.07 (4.22) & 10.89 (3.13) & 21.64 (5.60) & \textbf{60.35 (5.04)} \\
			& Force & 52.34 (15.51) & 36.01 (5.09) & 9.08 (3.56) & 22.74 (6.93) & \textbf{56.15 (10.45)} \\ \midrule
			
			\multirow{3}{*}{Segm.} & PCA & 6.02 (5.87) & -0.61 (5.26) & -11.66 (10.09) & -0.74 (10.87) & \textbf{16.00 (5.07)} \\
			& tSNE & 7.11 (2.95) & 0.02 (3.52) & -10.58 (7.63) & 5.54 (10.53) & \textbf{18.56 (4.49)} \\
			& Force & 6.35 (4.39) & -0.70 (3.68) & -15.48 (5.60) & 1.55 (7.16) & \textbf{17.12 (3.54)}   \\ \midrule
			
			\multirow{3}{*}{MNIST} & PCA & -24.39 (5.28) & 17.28 (4.71) & -6.78 (0.93) & -6.38 (1.31) & \textbf{25.14 (4.65)} \\
			& tSNE & -22.09 (9.28) & \textbf{19.26 (3.43)} & -6.05 (1.75) & -4.85 (2.00) & 13.42 (6.81)\\
			& Force & -28.30 (7.97) & 17.65 (4.68) & -6.13 (0.95) & -3.02 (2.85) & \textbf{22.07 (6.98)}
		\end{tabular}}
	\end{table}

Table~\ref{knerr} shows the neighbor error, which evaluates how far the 10 neighbors of each point in the visual space are in the original high-dimensional space. As an error metric, lower-values indicate better projections. For the neighbor error metric, the results are more diverse than the previous two metrics. While the proposed kernel version performs considerably better than the linear and baselines in the first two datasets (Wine and Cancer), it presents similar results with the linear and baselines in the other two datasets (Segmentation and MNIST). The neighbor error for the Cancer dataset (same setup as before) is visualized in Fig.~\ref{fig:cancer-err}. Note that the proposed method better captures the manifold structure of the data, creating smooth low neighbor error regions.

Overall, in all three evaluation metrics above, the kernel version achieves better results than the rest. However, the linear method outperforms the other two linear methods (LAMP, PLMP) in two out of four datasets (Wine, Segmentation). KELP is a kernel method so it would be fair to evaluated against our kernel method, where we perform better in every dataset and initialization combination.

\begin{table} 
	\centering
	\caption[Neighbor Error results]{Average Neighbor error after manipulation of control points. Best results marked bold. Standard deviation of 10 runs in parenthesis.}
	\label{knerr}
	\resizebox{1.0\textwidth}{!}{\begin{tabular}{ccccccc} \toprule
			\multirow{2}{*}{Data} & \multirow{2}{*}{Init} & \multicolumn{4}{c}{Neighbor Error} \\
			&  & \multirow{2}{*}{KELP} & \multirow{2}{*}{LAMP} & \multirow{2}{*}{PLMP} & \multicolumn{2}{c}{Proposed} \\
			&  &  &  &  & Linear & Kernel \\ \midrule \multirow{3}{*}{Wine} & PCA & 20.77 (5.84) &  26.24 (3.74)&  27.73 (4.01) & 24.63 (2.19) &  \textbf{8.33 (4.87)}\\
			& tSNE & 18.56 (3.41) & 27.75 (6.46) & 27.86 (4.42) & 25.92 (6.49) & \textbf{7.27 (2.34)} \\
			& Force & 19.03 (4.28)  &26.36 (5.39)  & 25.34 (2.68) & 25.81 (3.72)  & \textbf{12.52 (5.60)}  \\ \midrule
			\multirow{3}{*}{Cncr} & PCA & 12.37 (3.56) & 13.48 (1.48) & 15.52 (2.30) & 13.40 (2.90) & \textbf{5.23 (1.44)} \\
			& tSNE & 11.51 (2.39) & 13.58 (2.31) & 15.69 (2.47) & 14.48 (2.15) & \textbf{8.44 (1.78)} \\
			& Force & 12.43 (2.66) & 12.04 (1.66) & 13.62 (0.84) & 14.04 (2.23) & \textbf{7.49 (1.95)} \\ \midrule
			
			\multirow{3}{*}{Segm.} & PCA & 8.19 (1.38) & 7.00 (0.83) & 9.11 (1.12)& 8.83 (1.74) & \textbf{6.00 (0.83)} \\
			& tSNE & 8.52 (0.64) & \textbf{4.90 (0.65)} & 8.02 (0.95) & 7.51 (1.20) & 5.84 (0.77) \\
			& Force &7.69 (0.79) & \textbf{5.01 (0.90)} & 7.83 (1.47) & 7.27 (1.50) & 5.57 (0.97)   \\ \midrule
			
			\multirow{3}{*}{MNIST} & PCA & 50.19 (2.72) & 48.96 (3.20) & 46.31 (3.23) & \textbf{44.50 (4.10)} & 47.66 (2.88) \\
			& tSNE & 51.97 (3.20) & 50.60 (1.63) & 45.16 (4.94) & \textbf{45.10 (3.04)} & 48.03 (2.40)\\
			& Force & 50.15 (3.76) & 50.09 (3.34) & \textbf{46.57 (3.32)} & 46.87 (2.55) &47.91 (2.54)
		\end{tabular}}
	\end{table}

In general, during test time, the linear projection is very fast, requiring only the
calculation of one product between the projection matrix and an input sample. The kernel projection is more computationally demanding due to the calculation of the kernel function between an input sample and all the training samples. Nevertheless, due to our fast implementation in Theano, our linear and kernel versions achieve similar computational time. To illustrate that, we measured the fitting time of our method (time of gradient descent on cloning the initialization and the manipulation adjustment, or steps 2 \& 6 of Algorithm \ref{alg:iSIL_algorithm}) against the implementation of the baselines. 

\begin{figure}
\centering
\includegraphics[width=.45\textwidth]{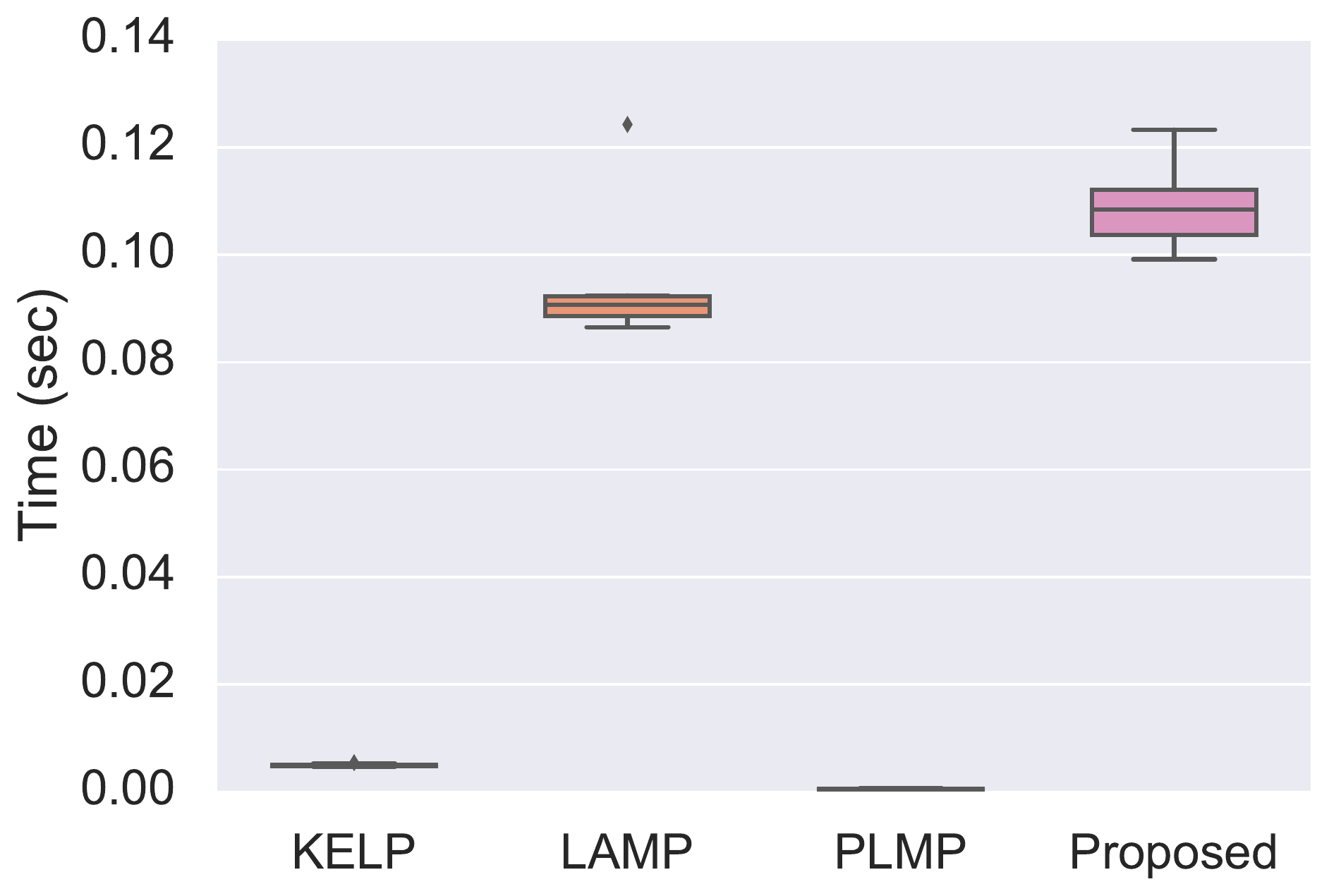}\quad
\includegraphics[width=.45\textwidth]{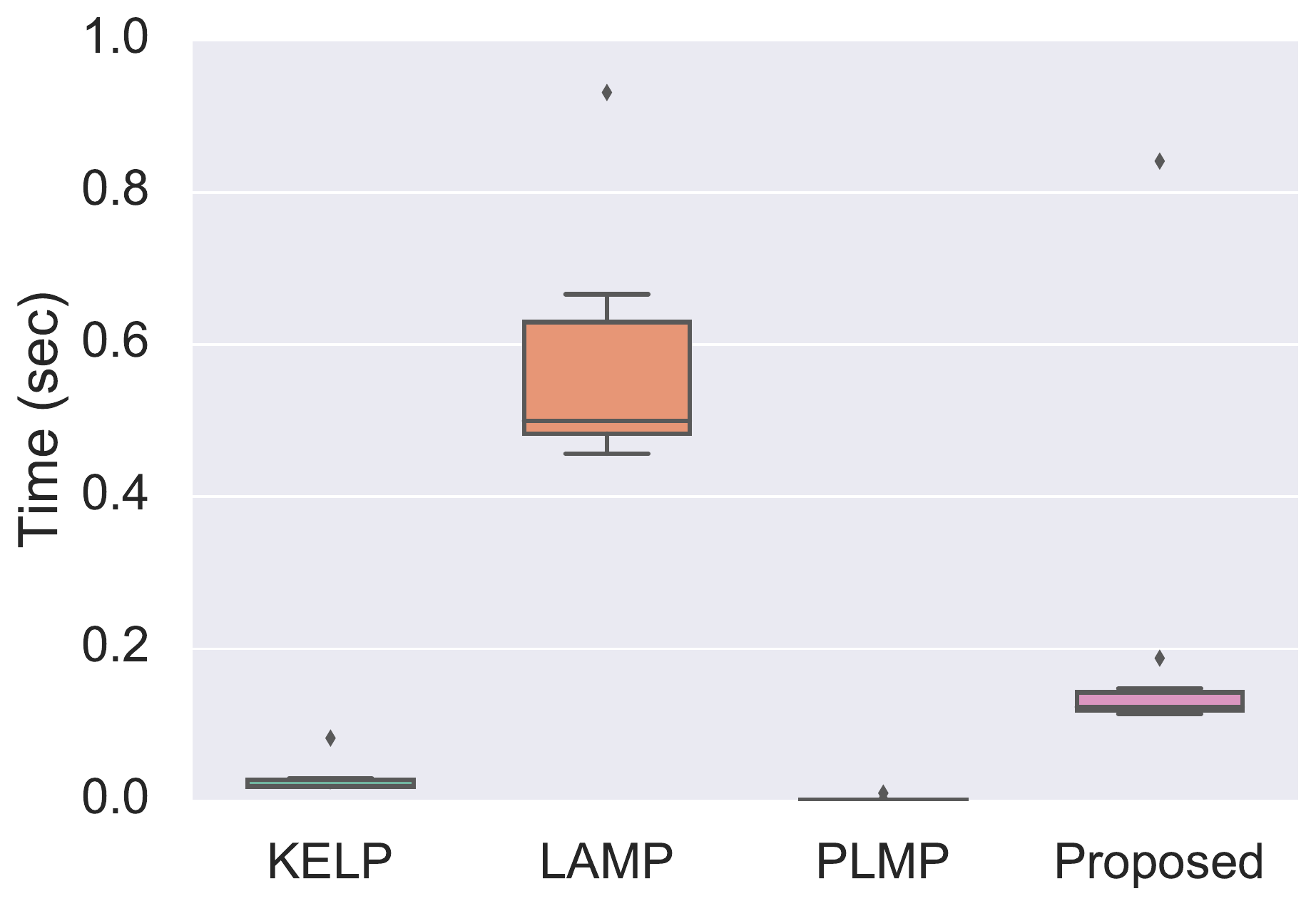}\quad
\medskip
\includegraphics[width=.45\textwidth]{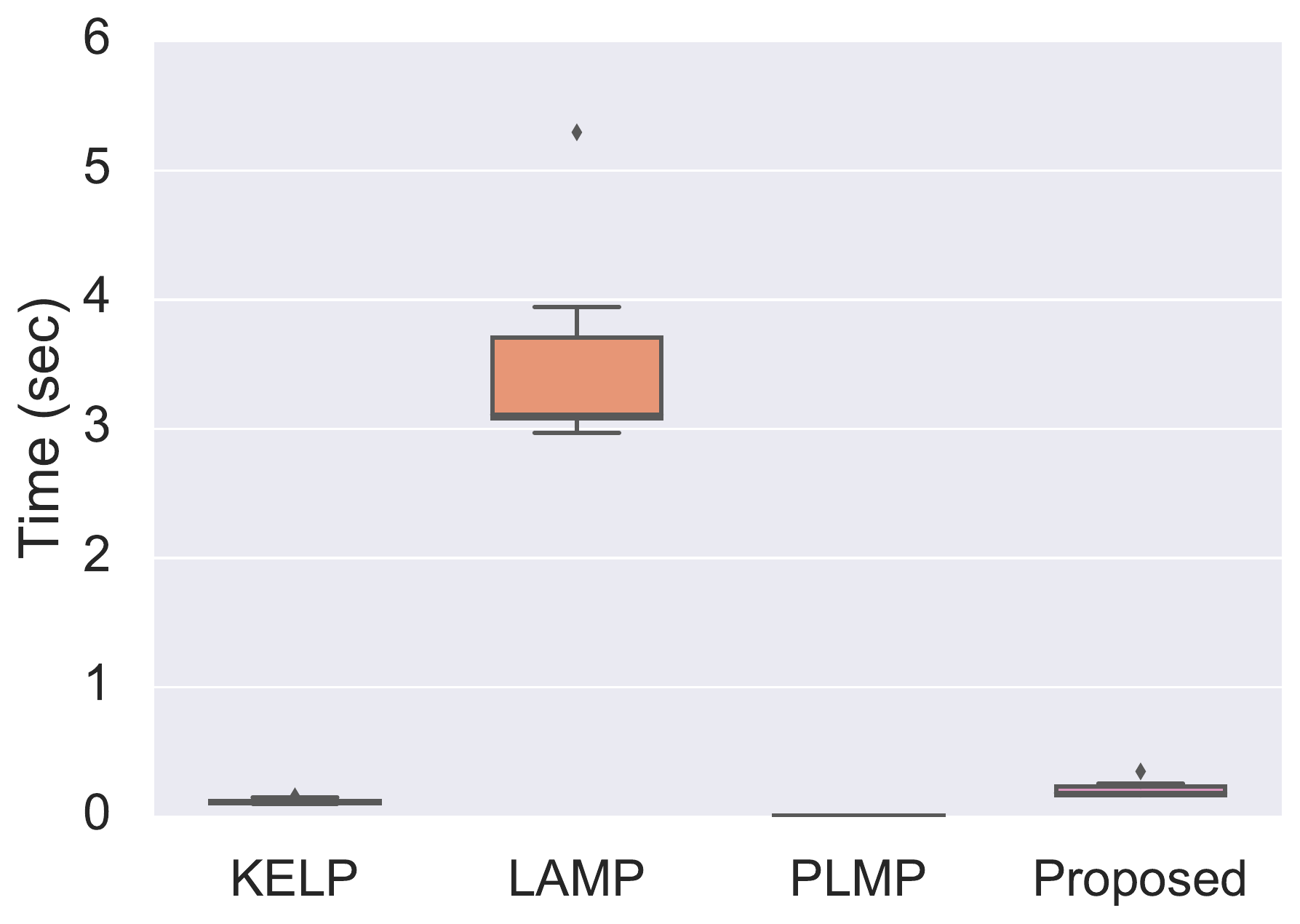}\quad
\includegraphics[width=.45\textwidth]{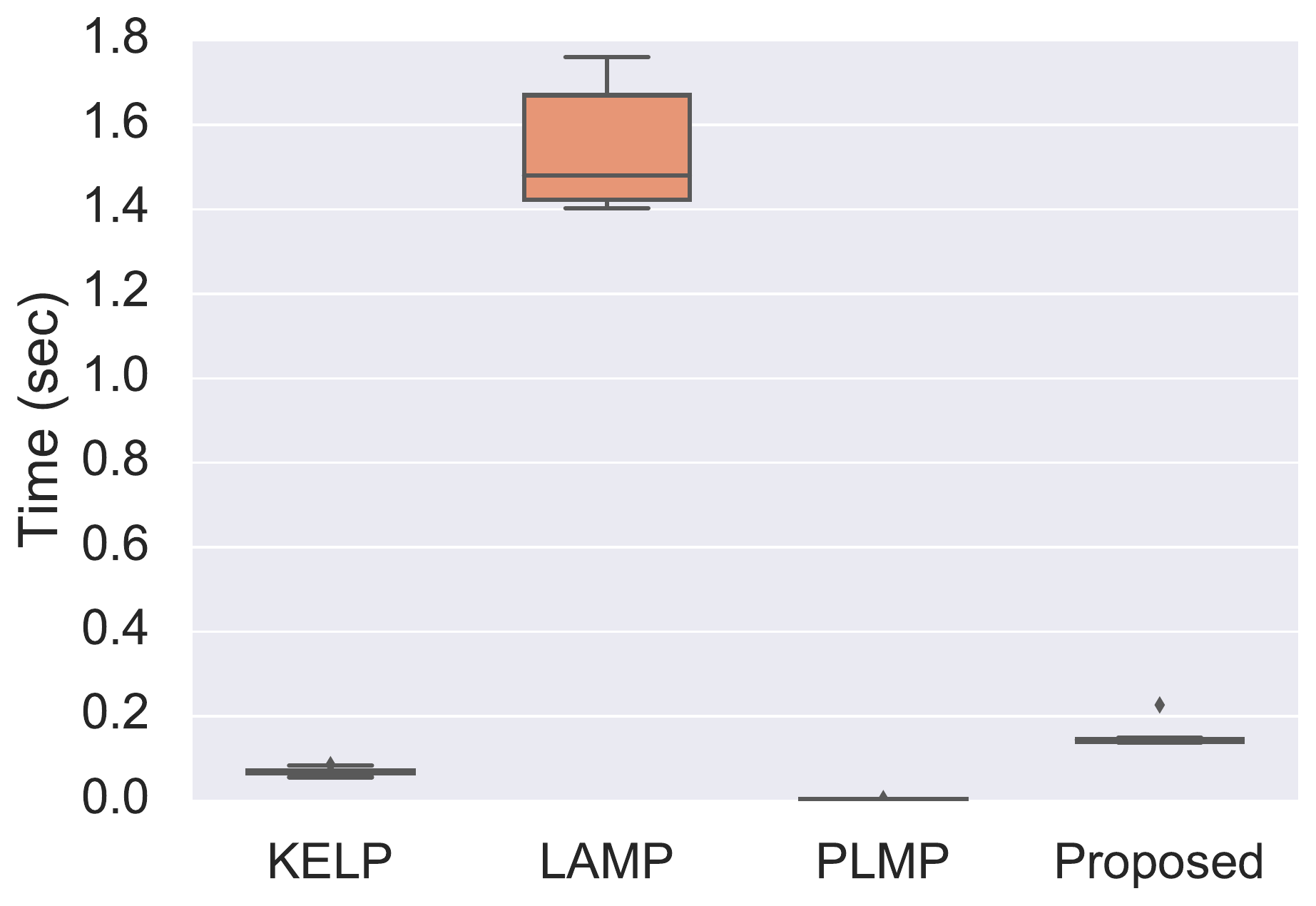}
\caption[Time comparison of baselines]{Time comparison of baselines on Wine (upper left), Cancer (upper right), Segmentation (down left) and MNIST (down right).}

\label{fig:time}
\end{figure}

In Figure \ref{fig:time}, we show that in small datasets like Wine our method is comparable to LAMP with around 0.10 seconds. In bigger datasets such as Segmentation and MNIST, LAMP appears to be significantly slower than the rest. PLMP is always the faster method, since it only requires the calculation of a couple of dot products, but it is the worst results-wise. Our proposed method achieves competitive times to PLMP and KELP, taking into consideration that the number of iterations is fixed to 500. By inspecting the loss minimization curve during the experiments though, we saw that our technique converges way before the 500th iteration most of the times. With that in mind, our computation time could be even lower. Please note that PCA was used as initialization while our linear method was employed. Qualitatively similar results came from the kernel method, as well as when tSNE and Force were used as initialization. All experiments were conducted on a MacBook Pro (Retina, 13-
inch, Early 2015) with a 2,7 GHz Intel Core i5 CPU and 8GB DD3 RAM.

\begin{figure} 
	\centering
	\includegraphics[width=1.0\textwidth]{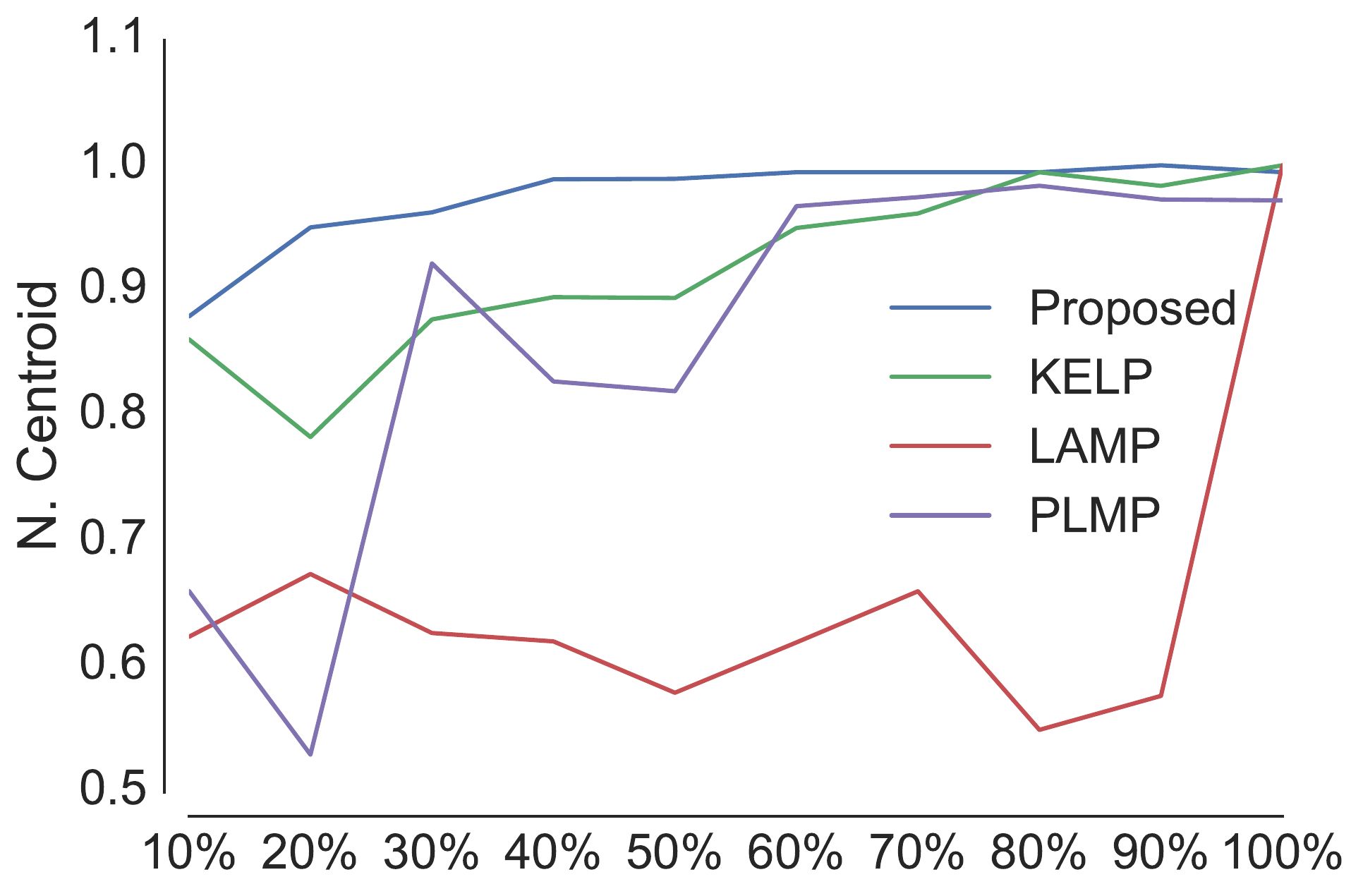}
	\caption[Control Point Percentage Change]{Nearest Centroid precision in relation to control point increase (over the baseline proposed in the literature, i.e., $\sqrt{(n)}$) on the Wine dataset.}
	\label{fig:percentage_wine}
\end{figure}

Finally, we evaluate the effect of the number of used control points on the quality of the learned projection. The results are shown in Fig.~\ref{fig:percentage_wine} using the proposed method and a linear projection function (PCA is used for the initialization). Even when a small number of control points is used, the proposed method outperforms the other methods. Qualitatively similar results were obtained using different initializations (e.g., tSNE or Force), as well as for other metrics and datasets.

\subsubsection{Interaction scenario \#2: Unveiling semantic manifolds via class dragging}

Apart from moving control points, interactive dimensionality reduction can provide us with new insights on datasets, such as discovering latent manifolds, or \textit{creating} new features and classes from existing data. In this section we test the second method described in the previous section, the neighbor-based embedding learning. 

The experimental procedure included the datasets of MNIST, Newsgroup and Head Pose due to their dimensionality and their sparse data structure (images and text). There is no comparison with the techniques used in the previous Subsection because they do not support such objectives beyond control points and interpolation. Unlike the previous where we move points towards their classes centers, now we are dragging a class or classes away. Please note that this task does not necessarily require labeled data; users can inspect the content (e.g images) and decide what they should move. Of course, inspecting images is an easier task due to being solely visual. For non-visual content such as text or even audio, visualization research is still open.

\textbf{Moving one class}. First, the interaction scenario of moving one class is going to be presented. A random subset of 500 digits of MNIST is selected, where these digits cover 4 classes (2, 4, 7 and 9). That class selection is not random since we want to exploit the morphological attributes of the shape of these digits. For example, some 4s are noted to be written like 9s and vice versa. Similarly, some 2s look like reverse 7s, when drawn with limited curvature. By dragging a class away, we expect that similar digits and classes will follow and will be positioned nearby. The evaluation of the interaction scenario is based mainly in visual--semantic observations, but also changes in Nearest Centroid are reported. In order to provide interpretability of our method, the trajectories of the positions before and after the manipulation are presented, with arrows denoting their beginning and arriving point.

\begin{figure}
\centering
\begin{subfigure}[b]{0.95\textwidth}
   \includegraphics[width=1\linewidth]{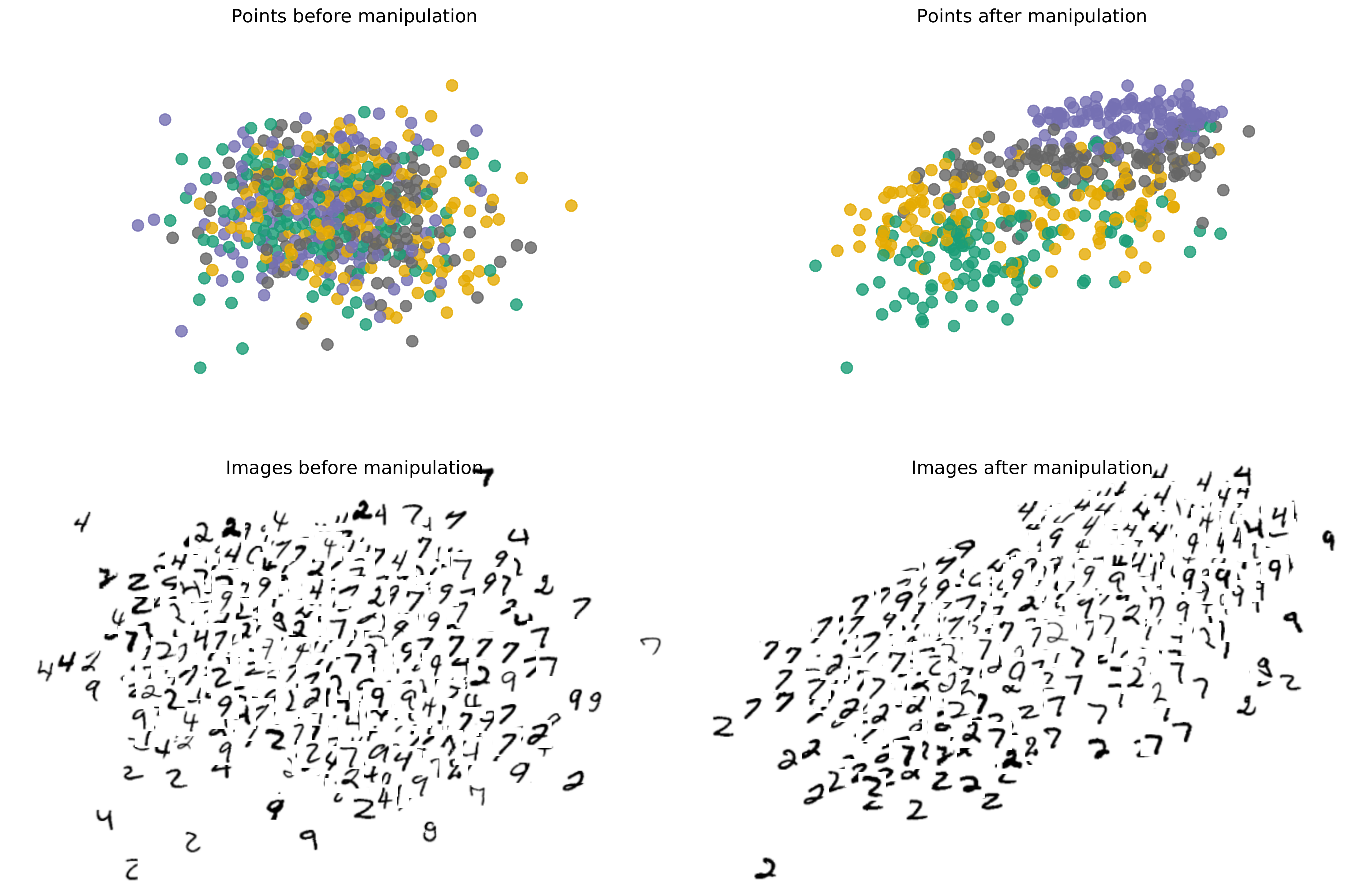}
   \caption{}
   \label{fig:mnist_linear}
\end{subfigure}

\begin{subfigure}[b]{0.95\textwidth}
   \includegraphics[width=1\linewidth]{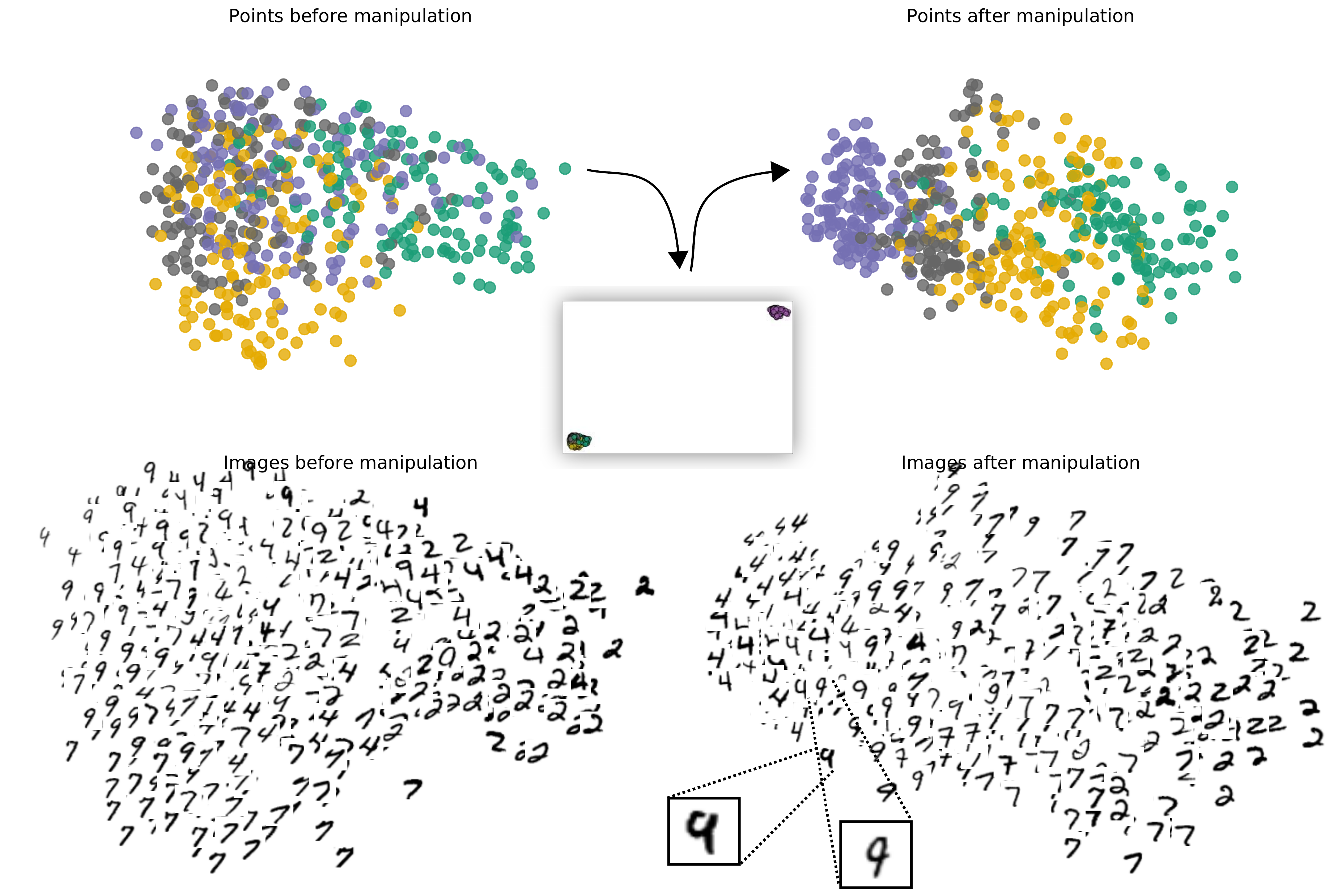}
   \caption{}
   \label{fig:mnist_kernel}
\end{subfigure}

\caption[Points of MNIST before, during and after manipulation along with their corresponding
images with some highlights, and trajectories]{Points of MNIST before, during and after manipulation along with their corresponding images and some highlights. (a) Linear, (b) Kernel version.}
\end{figure}

By using the Algorithm \ref{alg:iSNEL_algorithm}, we first load the dataset (500 digits of MNIST), clone PCA with our technique so that the initial projection is very similar to a PCA. The initial 784 dimensions are reduced to 68, keeping the 90\% of explained variance. Both the linear and kernel method are compared. In our kernel version, the initialization projection is also kernel, therefore the initial projection is essentially kernel PCA (learning rate $\eta=10^{-4}$). Then, the class of 4s is dragged away to the top right (+10 in x,y coordinates). The manipulated projection is fed into Algorithm \ref{alg:iSNEL_algorithm}, which stores the indices of the manipulated points. Then it finds the original and visual neighbors of such points and sets the target similarity to 1, while weighting the mask to 0.5 for visual neighbors, since we do not want them to be more significant from original ones. For this particular experiment, we choose 70 original  and 5 visual neighbors. Visual neighbors are set to 5 for all experiments with the 3 datasets. Afterwards, the difference of the new target matrix with the projection matrix is minimized with gradient descent for 500 iterations. Experimentation with learning rate and number of neighbors led to the reported results. In  Fig.~\ref{fig:mnist_kernel} the points before and after manipulation, the manipulation projection, the corresponding images of digits in 2D along with some illustrative examples, and finally their trajectories before and after.

By observing the initial (top left) kernel PCA projection in Fig \ref{fig:mnist_kernel}, we see that it does not manage to separate the 4 classes efficiently, neither to provide a meaningful manifold (middle left). On the other, by just dragging the class of 4s away (purple points), the similar 9s follow along (gray points), while most interestingly, the rest of classes are separated better. Indeed by moving the 4s away, we expect their most similar 9s to follow and the rest to be placed in a continuum covering 9s, 7s and 2s. As a side effect of the optimization we note that 2s are getting clustered, while 7s are still spread. Also near or inside the dragged class of 4s, we observe intruding 9s with very 4-like shape, so that even a human would confuse them. Two illustrative examples thereof are highlighted in the middle right position, the first one being a 9 with an incomplete top circle, and the second one being a probably single-stroke 9 with a prominent horizontal line, exactly like a 4.

Apart from visual evaluation of above manipulation, the Nearest Centroid precision is increased from 46.6\% in the initial projection to 74.5\% after manipulation (+27.9\%). Also the silhouette score follows a similar trend, from -0.018 to 13.5. But someone would argue that the original projection and the optimization are quite good because of the power of kernel method. The outcome of the same manipulation via the linear method, so that the initial projection is essentially a simple PCA, and the new projection a fast linear mapping is show in Fig.~\ref{fig:mnist_linear}.

\begin{figure} 
\centering
\includegraphics[width=\textwidth]{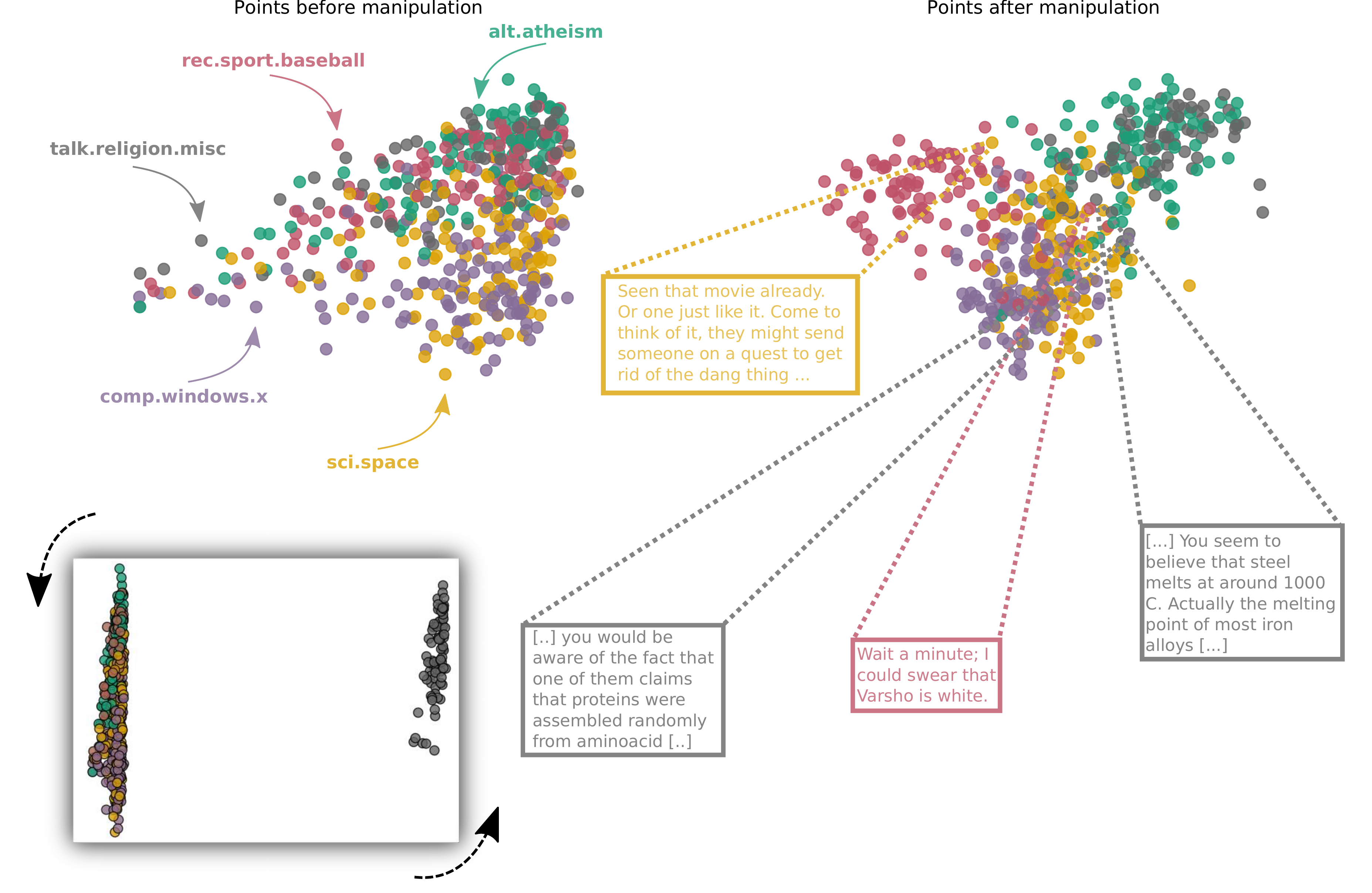}
\caption[Neighbor-based Manipulation on Newsgroup - Kernel Version]{Points of Newsgroup before and after manipulation along with some highlighted documents. Kernel version.}
\label{fig:newsgroup_kernel}
\end{figure}

Indeed, even with the linear version in Fig \ref{fig:mnist_linear}, the result is probably more dramatic than before, because the initial projection is worse. The initial PCA (top left) does not manage to capture any significant structure of data and looks like a random projection. On the other hand, the manipulation managed to unveil the  same continuum of classes described above, so that 4s are followed by 9s, 7s and 2s. Nearest Centroid Precision before manipulation (30.5\%) is expectedly low, rising to 59.5\% after (+29\%). On absolute precision terms, the kernel version earlier achieved a higher precision (74.5\%), but the relative increase of the linear version is marginally higher (+29\% vs +27.9\%). 

Following the same interaction scenario (drag one class away), we choose the \textit{religion} class of Newsgroup dataset and drag it horizontally to the right. With that interaction, we expect the \textit{atheism} class to follow along, since their content might overlap and the rest of classes (\textit{space, baseball and computers} to auto-organize as earlier. Also, we expect \textit{space} to be closer to \textit{religion} and \textit{atheism} since they include more intellectual topics, while \textit{computers} and \textit{baseball} to be further away. Bellow is the kernel version of the neighbor-based manipulation of a subset of 500 documents in Newsgroup, setting the original neighbors to 20. Classes are annotated according to their color, while some illustrating documents are highlighted.

The neighbor-based interaction (Fig \ref{fig:newsgroup_kernel}) seems to be effective on the Newsgroup dataset, validating the assumptions that the \textit{religion} class overlaps with \textit{atheism}. Even though the initial projection is kernel PCA, it does not succeed in separating the classes well enough (39.1\%), whereas after the interaction the rest of classes are  more compact (60.3\%). For example, \textit{baseball} class (red) is unlikely to be high-dimensional neighbors with \textit{religion}, but appears to have been impacted by the optimization, getting clustered top left.

In case of visualizing text, most available methods include word clouds or word trees, but these methods struggle to offer reasonable representations beyond the word level. In our case, we simply cannot display all documents in the projection, so we choose some illustrating examples thereof.  An interactive scatter plot with document annotations on hover is going to be built as future work. Regarding the highlighted documents, by inspecting some outliers of residing in the middle of the four classes, we gain insights on where our method placed each document and why. For example, the top right highlighted document in Fig. \ref{fig:newsgroup_kernel}, is a \textit{religion} document placed very close to the \textit{space} class. If we inspect its content, this document talks about the melting point of iron and temperatures: 

\begin{displayquote}
\textit{"...even if the gun was found in area which achieved the 1000 C temperature, the steel parts of the gun would not be deformed, and it would still be trivial to identify the nature of the weapon. A fire is not an isothermal process..."}
\end{displayquote}

Our method correctly placed it near the science class. For sanity check, this is a random document of the \textit{religion} class, inside the gray cluster:

\begin{displayquote}
\textit{"...if you decide to follow Jesus, of which I indeed would be ecstatic, then all the glory be to God..."}
\end{displayquote}

On a similar example, the red highlight in Fig. \ref{fig:newsgroup_kernel} belongs to the \textit{baseball} class, but it is placed near the \textit{religion, atheism} cluster, probably due to the occurrence of the word "swear". Again, for sanity check, this is a random document of the \textit{baseball} class, inside the red cluster:

\begin{displayquote}
\textit{"...Yes, Ivan Rodriguez, last year.  Batted .260 and threw out 51\% of the baserunners. Not too shabby for a rookie from AA. 20 years old last year..."}
\end{displayquote}

Apart from this qualitative evaluation of the placement of semantically similar documents nearby, the Nearest Centroid precision before (39.1\%) and after (60.3\%) interaction, increases by +21.2\%. Silhouette clustering coefficient follows the same trend with before (-2.9) and after (7.6).

\begin{figure*} 
\centering
\includegraphics[width=\textwidth]{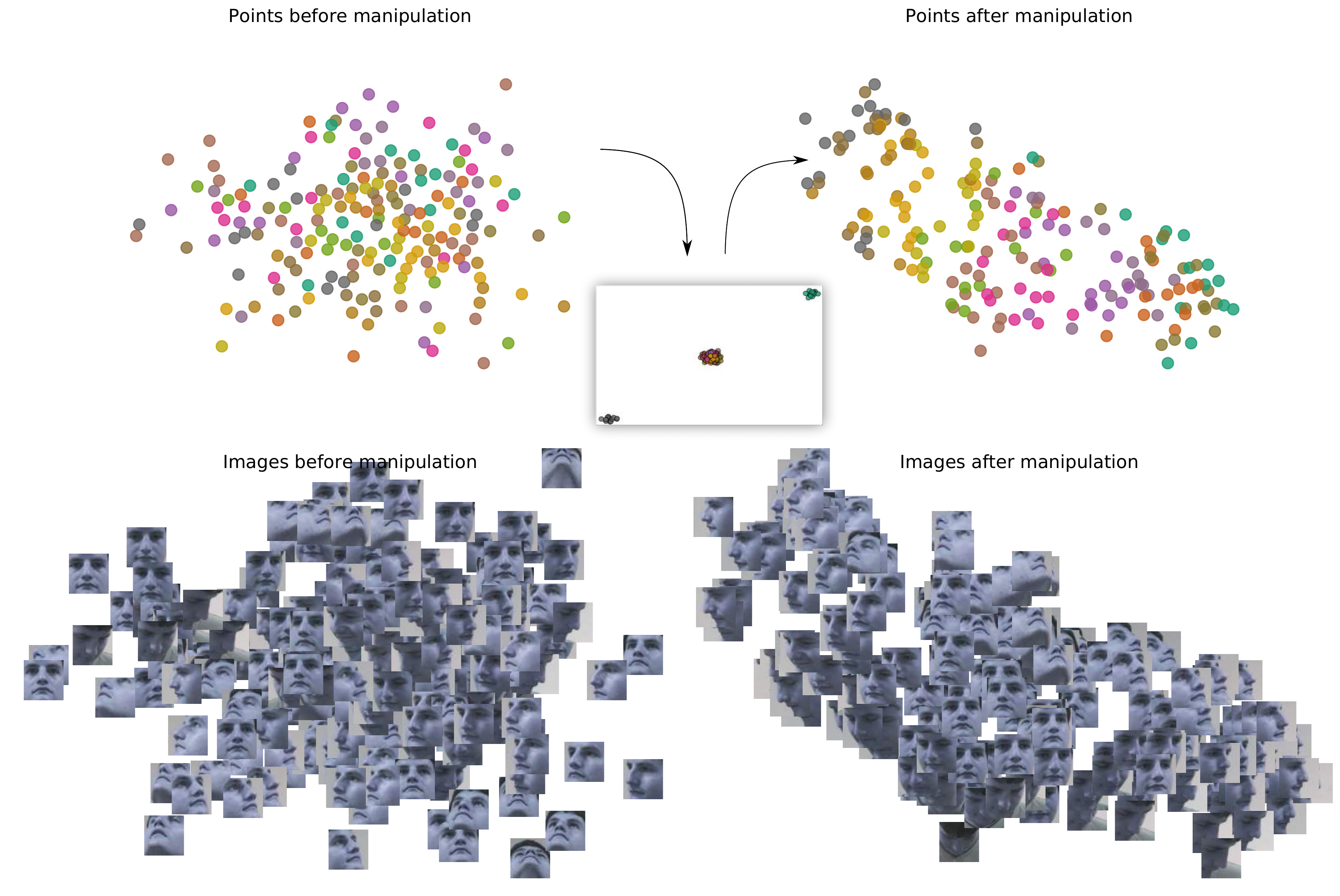}
\caption[Neighbor-based Manipulation on Head Pose - Linear Version]{Points of Head Pose before and after manipulation along with their corresponding images. Initialized with PCA. Linear version.}
\label{fig:face_linear}
\end{figure*}

\textbf{Moving multiple classes}. On a slightly different interaction scenario, two classes of Head Pose dataset are moved afar, so that the class corresponding to -90 angle pan goes down left, while the one corresponding to +90 pan goes up right. That way, the assumption is that our method will \textit{unroll} the faces manifold which covers the range from -90 to 90. Please note that of the two available class variables (tilt and pan), only the pan class is taken into account in this interaction. The reasoning behind that choice is that even by moving those faces that turn their head horizontally to the two sides, our method will place the tilted faces vertically on the $y$ axis. A subset of Head Pose dataset is used; all 186 images of a middle aged man. 29 dimensions are kept after PCA. The outcome of manipulation is shown in Fig.~\ref{fig:face_linear}.

By dragging the top left and right angle of faces (Fig. \ref{fig:face_linear}), indeed the manifold unrolls so that every in-between angle is placed in a continuum. But the most interesting  finding is that the faces unroll on the \textit{vertical axis} as well, as a side effect of the optimization. Due to smaller dataset, original neighbors are set to 40. In more quantitative terms, the initial PCA struggles to separate the classes (16.4\% precision), whereas the manipulation resulted in +26.4\% increase (42.8\%). Kernel version produced qualitatively similar results.

\begin{figure}
\centering
\includegraphics[width=1.0\textwidth]{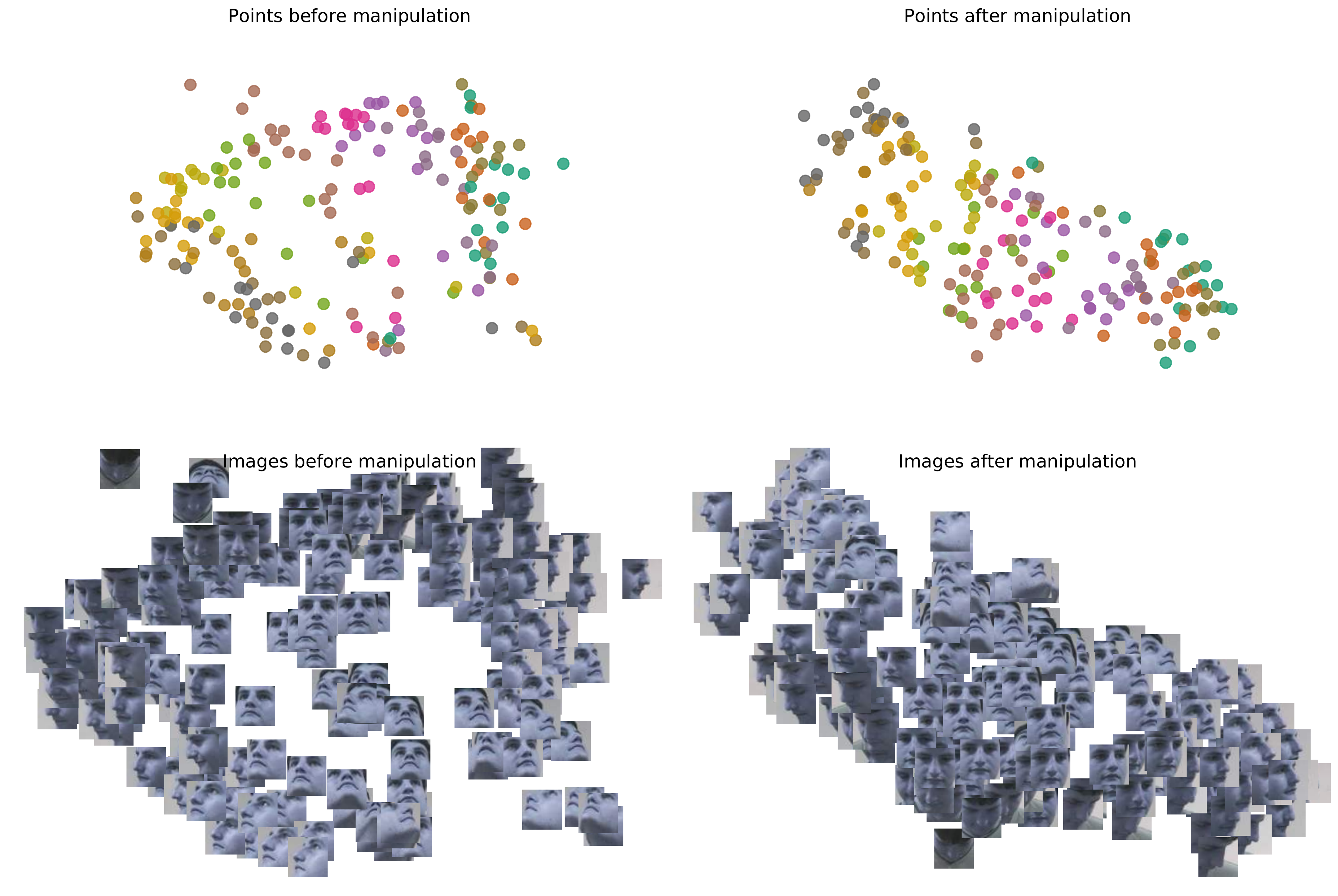}
\caption[Neighbor-based Manipulation on Head Pose - Linear Version - tSNE]{Points of Head Pose before and after manipulation along with their corresponding images. Initialized with t-SNE. Linear version.}
\label{fig:face_linear_tsne}
\end{figure}

Someone would argue that since the original paper of Isomap faces \citep{tenenbaum2000global}, there have been a lot of algorithms that can be used as an initialization instead of PCA, like Isomap, LLE or t-SNE. These algorithms have been proved to capture semantic manifolds in image data \citep{maaten2008visualizing}. Therefore, instead of PCA, the same experiment as in Fig \ref{fig:face_linear} was repeated by cloning t-SNE as a first step. The outcome is shown in Fig.~\ref{fig:face_linear_tsne}.

Even though t-SNE uncovers a semantic manifold of faces in Fig. \ref{fig:face_linear_tsne}, it struggles to separate the tilt aspect (vertical angle) of faces, while it does not succeed always with the horizontal angle. For example in the top left it placed nearby both upwards facing and downwards facing faces. On the other hand, our manipulation again unrolls the semantic manifold of faces in the horizontal and vertical axis. Precision increase might not be so dramatic (+9\% for our method), but as discussed earlier, the precision metric for the specific dataset is somewhat misleading. For instance, some gray points might be far in the visual space but they represent the vertical angle.

\section{Discussion and Conclusions}
For our experiments, different interaction scenarios were presented during which we showed that iSP is outperforming competitive techniques of multi-dimensional projection while they offer more interaction capabilities in order to explore high-dimensional data.

\subsection{Results}

By using this similarity --rather than distance-- approach, we observe that our linear and kernel versions outperform competitive baselines in benchmark dataset. Namely, our methods improve classification precision in Wine dataset by 16--29\%, in Cancer by 1--31\%, in Segmentation by 2--25\% and in MNIST by 4--48\%. Similarly, our methods improve the clustering coefficient in Wine dataset by 12--43, in Cancer by 5--51, in Segmentation by 11--30 and in MNIST by 6--53. Also, our methods reduce stress and are able to run in competitive times. In short, we propose an end-to-end framework for control points-based interpolation, relying on optimizing similarity matrices and manages to outperform baselines. On top of that, the code of the prototype interactive scatter plot is going to be provided as open source, in order to enable more analysts to perform such tasks.

Apart from evaluating our methods with traditional classification and clustering metrics, inspired from multidimensional visualization research \citep{lespinats2011checkviz}, alternative visual evaluations were explored. The metric of Neighbor Error was formulated so that it can evaluate how far the neighbors in visual space are in the high-dimensional space. Again, our methods reduce neighbor error in Wine dataset by 10--20\%, in Cancer by 6--10\%, in Segmentation by 0--3\%, and in MNIST by 1--12\%. An observation arising from the Neighbor Error plots is that our method tends to place points with the lower neighbor error \textit{outside} the area where more clusters reside. Also by seeing the trajectories of points before and after the manipulation in image and videos, we gain a better insight on how the iSP transforms the topology iteratively. In short, by exploring alternative visualizations, we are able to understand multi-dimensional projections better.

Regarding the second interaction scenario, that of dragging classes away, there are some interesting observations arising from the experiments. First, we saw that it is possible to perform better separation in classes just by dragging a class away. Please note that no class information was taken into account in any experiment. Indeed, Nearest Centroid proves that by interaction we achieve better classification precision in MNIST by 29\%, in Newsgroup by 21\%, and in Head Pose by 26\%. But the precision increase tells half the story here, since the most surprising results come from the semantic manifolds uncovered by dragging classes away. By moving the 4s in MNIST, we observe a complete reorganization of the rest classes, being placed in a continuum according to their shape. In text documents of Newsgroup, we observe which documents overlap semantically but also which classes are the odd-one-out (\textit{sports} being placed in a more compact cluster than \textit{religion, atheism, science and technology}). Second, by comparing our method with sophisticated, state-of-art non-linear dimensionality reduction techniques, we show that interaction unveils the hidden structure of data in a more meaningful way. In short, interaction helps discovering new, latent semantic manifolds in datasets. 



\subsection{Future work}

During our experiments, some interesting properties of the supervised method came up. For example, by cloning the LDA method so that we simply set the similarity of points belonging to the same class to 0.8, and by training the kernel version for many iterations (over 1000), we observe that the points tend to be placed in almost perfect spheres in the visual space. While this procedure alters the distribution of the dataset, it could be used in order to \textit{force} manifold structure in datasets that do not present such attributes.

\begin{figure} 
\centering
\includegraphics[width=1.0\textwidth]{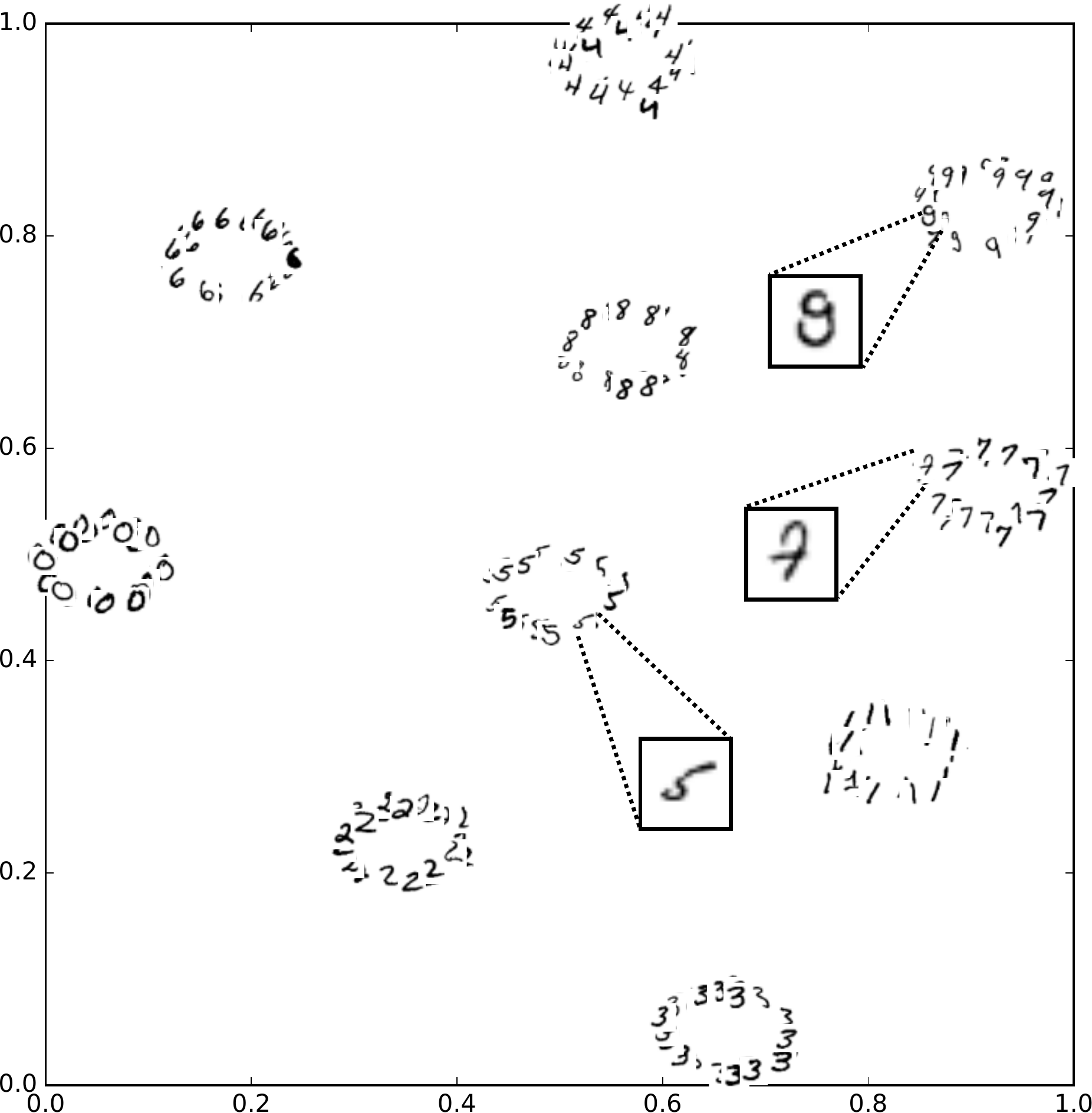}

\caption[2D projection of MNIST - Circles emerging]{Circles emerging by over-training the LDA-like kernel version on MNIST.}
\label{fig:mnist_2D}
\end{figure}

This means that when traditional dimensionality reduction methods like t-SNE or PCA struggle to unveil a meaningful structure of data, this above trick can force data to be placed in semantic manifolds. Someone would argue that these are just digits of the same class clustered together, but if we inspect where each class is placed in space, we discover that they share semantic attributes. For example, in Fig.~\ref{fig:mnist_2D} 300 digits of MNIST trained with the same objective for 2500 iterations are shown.

Apart from perfect classification and clustering in 2D, we observe some interesting attributes in Fig. \ref{fig:mnist_2D}. First, similar classes (e.g 4s and 9s, 1s and 7s) are placed nearby, while different ones (0s, 6s) are considerably far. Second, if we inspect where each point resides in their circle, we discover that they are not placed in random. For instance, the highlighted 9's shape above resembles a regular 8 a lot, that is why it is placed in the most 8-wards facing position in its class. Same applies to the highlighted 7, which is the only one drawn with an almost circular top, like an 8 or 9. Also, the highlighted 5's oblique shape looks like an 1 on its side. 

Recently, a very interesting paper provided interactive capabilities to tSNE \citep{pezzotti2016approximated}, but the authors focused mostly on approximating the computationally intensive task of nearest neighbors calculation for scalability, while the data manipulation part included scenarios conceptually different than ours such as inserting, deleting data-points and dimensions. Nevertheless, we consider these interaction scenarios and their proposed method as future work, taking into account that there is no available official implementation yet.

It should be stressed again that the iSP allows to easily derive new techniques. The target similarity matrix can be set according to another multi-dimensional projection technique, like the KELP, in order to provide fast linear projections. Essentially, it can clone the projection of every other projection technique and offer out-of-sample extensions. Preliminary experiments show interesting results. Except for cloning techniques, more complex targets can be set, such as the distribution produced by the output of a deep convolutional network. Also the loss function used in this work could be also used to back-propagate the error through the layers of a deep neural network. We are also considering an online-batching version of our method so that users can make modifications on the iterations of the gradient descent, towards progressive visual analytics. 

Inspired by the interesting manifolds arising from interacting with sparse data, we consider a couple of different applications for this interaction scenario. For example, most Natural Language Processing tasks today depend on dense word embeddings to represent text. Usually, pre-trained models of huge train-sets are used as off-the-shelf feature extractors for such tasks. These pre-trained models are trained in generic datasets such as Wikipedia or Google News that are not tailored to specific tasks (e.g., sentiment). A possible solution is to train from scratch our specific dataset with the word2vec method. However, this requires a lot of time, computation and data. In case we want to use above models as \textit{tailored} feature extractors, our interaction methods could be used so that we adapt embeddings according to our task. A similar application could be in the field of computer vision, where pre-trained generic models, such as VGG \citep{simonyan2014very}, could be adapted to specific tasks, through interaction.

\bibliography{example} 
\bibliographystyle{IEEEtran}

%




\end{document}